\documentclass{IET}

\begin{document}

\title{Visual saliency detection: a Kalman filter based approach }

\author[1,*]{Sourya Roy}
\affil{Department of Instrumentation and Electronics Engineering,Jadavpur University, Kolkata-700032, India}

\author[2]{Pabitra Mitra}
\affil{Department of Computer Science and Engineering ,Indian Institute of Technology, Kharagpur - 721302, India}

\affil[*]{souroy099@gmail.com}

\abstract{In this paper we propose a Kalman filter aided saliency detection model which is based on the conjecture that salient regions are considerably different from our "visual expectation" or they are "visually surprising" in nature. In this work, we have structured our model with an immediate objective to predict saliency in static images. However, the proposed model can be easily extended for space-time saliency prediction. Our approach was evaluated using two publicly available benchmark data sets and results have been compared with other existing saliency models. The results clearly illustrate the superior performance of the proposed model over other approaches.}

\maketitle

\section{Introduction}\label{sec1}
\label{sec:introduction}
Saccadic eye movement is one of most significant feature of human visual system which helps us to scan a scene with incredible celerity and robustness. During saccadic eye movement human eyes rapidly moves from one point to another while simultaneously detecting interesting regions. Modelling and automatic detection of these salient regions which essentially seek attention of human visual system, is currently a problem of considerable interest. It should be apparent that early detection of salient regions have numerous important applications. From scene understanding to rapid target detection, more or less every computer vision task can be aided by saliency prediction.
Previous approaches for saliency mapping can be divided into two groups: bottom up and top down. Bottom up approaches relies on processing of inherent features (like contrast, edges etc.)of the image and do not depend on any priori information, while top down hierarchy inspired methods assume that past experience and knowledge plays an important role in driving attention. 
\begin{figure}[t]
\begin{center}$
\begin{array}{ccc}
\includegraphics[width=1in]{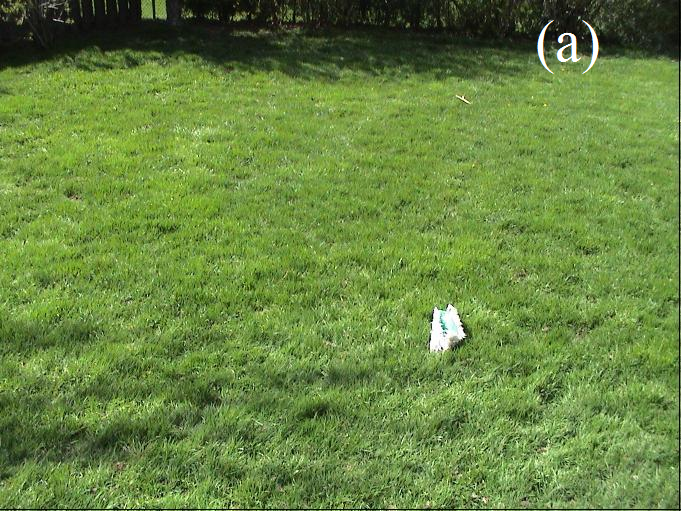}\hspace{-0.1em}
\label{fig:Stupendous}   &
\includegraphics[width=1in]{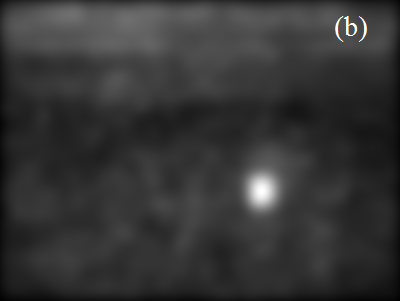}\hspace{-0.1em} &\vspace{.6em}\\ 
\includegraphics[width=1in]{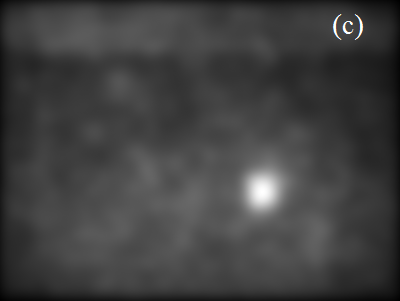}\hspace{-0.1em}&
\includegraphics[width=1in]{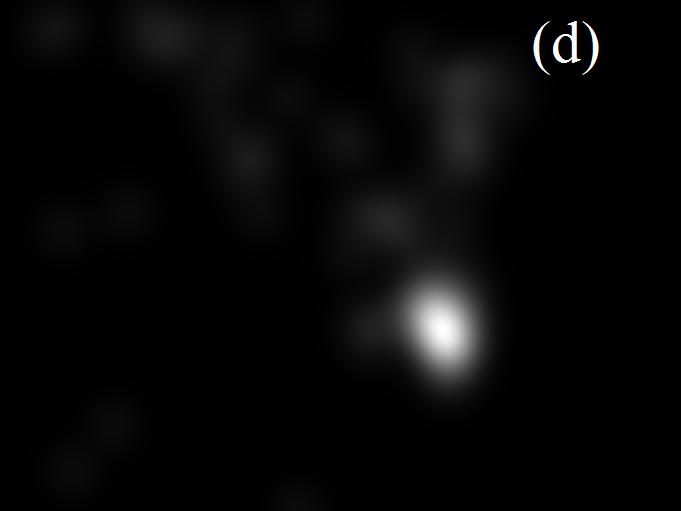}\hspace{-0.1em}

\end{array}$
\end{center}
\vspace{-0.7em}
\caption{(a)Input image, (b) Saliency map using proposed method (with 3 feature channels) ,(c)Saliency map using proposed method (with 7 feature channels), (d) Human fixation density map  }
\end{figure}

Bottom up saliency detection methods are generally termed as low level methods as they mostly utilizes low level features of the image. These group of methods can be further classified into biologically inspired approaches, purely computational techniques or methods which lies more or less in the middle-ground. Inspired from the “feature integration theory” ~\cite{Treisman198097}.  the saliency model proposed by Itti et al ~\cite{Ittikoch}  is undoubtedly the most influential and significant work from the first category. This biologically inspired model segregates the input image into several simple feature maps and calculates center-surround difference for each map and finally combines them in a linear manner to produce the master saliency map.   Bruce and Tsotsos ~\cite{Bruce}  proposed a saliency model which is based on self-information maximization of any region relative to its surrounding. . Seo and Milanfar~\cite{Seo} also compared the center and surround using a “self-resemblance” measure.Murray et. al~\cite{Murray}   proposed a method which utilizes a low level vision model of color perception. Zhang et al ~\cite{Zhang} computed saliency as a probability of target availability in a region. In 2005, Itti and Baldi~\cite{Itti2} proposed a “Bayesian–surprise” based method defines saliency as a measure of “surprise”.  Completely or partially computational approaches are also common in past literatures . 

In this paper we propose a kalman filter based saliency detection mechanism which is motivated by two much-discussed biological phenomena: 1)  deviation from visual expectation or visual surprise~\cite{Expct1}~\cite{Expct2}~\cite{Itti2}~\cite{Itti3} and saccadic eye movement~\cite{Scd1}~\cite{Scd2}. Our algorithm does share the generalized notion of ‘surprise’ presented in a previous work~\cite{Itti2} ~\cite{Itti3} by Itti and Baldi where they have proposed a method for video saliency detection, however using one of the most commonly used estimation technique i.e. Kalman filter, we have developed a more compact and flexible method for calculating ‘surprise’ based saliency in a static image and our model can be easily extended for video case. Our algorithm has three main stages. First the color input image is split into low-level visual feature channels. Based on the choice of feature channels, we have implemented two variants of our model. The first one uses only three opponent color channels and the other one uses seven features channels [as in Itti-Koch model]. Then for each channel individual saliency map is generated using Kalman filter algorithm and lastly all of them are combined to produce the final map. To employ kalman filter model for saliency mapping we have assumed that the input image is a noise corrupted measurement (perceived) signal and the true signal is an “expected image” which our visual system generates. So to produce the saliency map corresponding to a feature channel, we first estimate the expected counterpart of that specific channel using kalman filter. Then we simply calculate the difference between the expected and perceived feature channel and define saliency as the magnitude of difference. The main contributions of our work are as follows:

1) A definition of ‘visual surprise’ in static image. 

2) A bottom-up saliency model based on Kalman filter. 

3) Evaluation of the proposed model on two popular benchmark data sets. 

\section{\uppercase{Kalman filter based saliency detection}}
\begin{figure*}[t]
\begin{center}

\includegraphics[width=0.9\linewidth]
                   {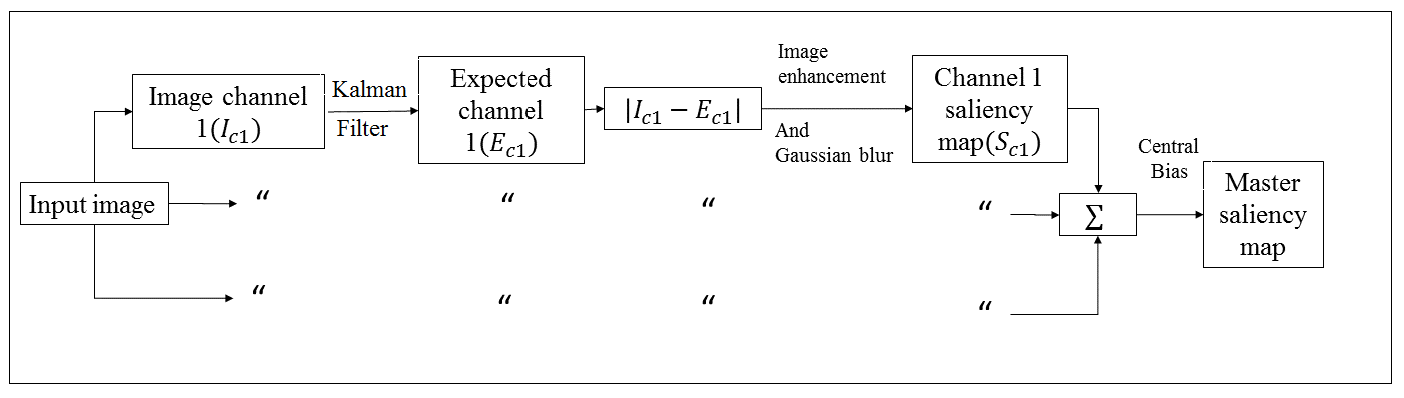}
\end{center}
   \caption{Flow diagram of the proposed model. This diagram represents the first variant of the proposed model which initially splits the input image into three opponent color maps and uses them as feature channel for saliency detection. The second variant follows exactly same framework as this, but uses seven feature channels (one intensity, two color and four orientation channels).}
\label{model_f}
\end{figure*}

In this section we will describe our model thoroughly along with details of implementation. The basic architecture of the proposed algorithm has been shown in Fig. \ref{model_f}.
\subsection{Definition of visual surprise in static image}
When humans (also many other animals) look at a scene, their eyes move and scan the scene in a rapid jerk like motion and this is called ‘saccadic eye movement’. During two rapid saccades, eyes stop at fixation points. Naturally, these fixation points indicate the salient regions in a scene. Now Itti and Baldi~\cite{Itti2}~\cite{Itti3} and also others~\cite{Expct1}  already showed in their work that our visual system encounters surprise in these regions and our visual prediction(based on prior belief) will be more different from the actual input. In their work Itti \& Baldi~\cite{Itti3} dealt with video data where pre and post frame can be treated as prior and posterior data respectively. 
So visual surprise can be easily computed by calculating how much the posterior is dfferent from the prior. But, unlike video data, there is no pre or post frame in a single static image, to tackle this problem, we will move from one randomly selected block of an image to another while treating the former as prior (${\omega}_{k}$) and the later block as posterior (${\omega}_{k+1}$). However, we don't compare the blocks to calculate visual suprise. Instead of that,we simulate a process where we learn an unknown relation between the prior space (${\omega}_{k}$) and its local statistics, then using that relation we are trying to estimate the next region or the posterior,   (${\omega}_{k+1}$). So visual surprise of any particular pixel can be defined as: 
\begin{equation}
\text{Surprise}=|\text{Estimated value}-\text{Actual input value}|
\end{equation}
We will term the entire esimated image as "visually expected image". In the next section, we have presented formal definitions of  "visually expected image" and it's corresponding "saliency map". 

When modelling visual surprise, we have also considered an intuitive hypothesis in our work, that is when we are encountering more than a certain level of error in prediction, we become more ‘visually aware’ and rely less on our prior belief; vice versa occurs when perceived input image is continuously agreeing with our expectation. In our model 'visual awareness' decides to which extent our  expectation gets modified by the posterior data. We can relate this intuitive idea with our daily experiences, for example: if a car comes suddenly in front of us when we are walking on a road, we will be surprised and for the rest of the path we will stay more cautious. Also it have been assumed that when we shift our eyes to a distant part of scene, our reliance in prior belief reduces. So both distance and error in prediction controls the trade-off between visual awareness and reliance in prior.  In section 2.3 we will describe how this scheme can be implemented by manipulating two design parameters of kalman filter.  


\subsection{Definition of expected image and image saliency}
As we have already discussed that the heart of our algorithm is the generation of the “visually expected image”. Now to generate visually expected image, ${E}_{c}$ corresponding to the input image channel, say ${I}_{c}$, we have used a coarse construction policy i.e. the expected image will be coarse in nature. To simulate this we have assumed that ${E}_{c}$ will be consist of equally sized regions (in our case: blocks of dimension $m\times n$ ) and all pixels of any specific region/block will have same value. So each uniform block can be defined by only one value. Let’s say, ${M}_{k}$ is the value of all pixels in the kth block of ${E}_{c}$ and  its channel input counterpart is the kth block , ${\omega}_{k}$(i.e. $\omega_{k}\subseteq I_{c}$ ). Now ${M}_{k}$  is defined as follows: 

\begin{equation}
\begin{split}
{M}_{k}=\sum_{i=1}^{3}{e}_{ik}.\text{Local entropy}_{scale_{i}}+\sum_{i=1}^{2}{m}_{ik}.\text{Local mean}_{scale_{i}}+\sum_{i=1}^{2}{s}_{ik}.\text{Local standard deviation}_{scale_{i}}
\end{split}
\end{equation}

Where ${M}_{k}$  is a linear combination of  local statistics. Our first task is to estimate the values of the coefficients (i.e.   ${e}_{i}$ ,${m}_{i}$, ${s}_{i}$) of  ${M}_{k}$  and using that we will construct the coarse expected image, ${E}_{c}$. For this estimation purpose we have used Kalman filter. After constructing expected image ${E}_{c}$ associated with ${I}_{c}$, the saliency map Sc  corresponding to ${I}_{c}$, can be computed as follows:
\begin{equation}
S_{c}=|I_{c}-E_{c}|
\end{equation}

After computing saliency map for each channel we combine them and apply center bias to generate the final saliency map. However, before combining these channel saliency maps, we enhance them individually via ‘contrast stretching’ to make the salient regions more conspicuous. 

Though we have defined  ${M}_{k}$ as a linear combination of statistics, our model doesn't impose any restriction on the choice of  ${M}_{k}$.  ${M}_{k}$ could have been more simpler or more thoughtfully crafted nonlinear combination of features. 
%


\subsection{Kalman filter algorithm}
In our work we have assumed that the input image is the measurement signal and the predicted image by our visual system is the true signal. So using the Kalman filter algorithm~\cite{Kal1}~\cite{Kal2}~\cite{Kal3} we will estimate the true signal using block to block random traversal policy which has been already described in previous section. Kalman filter is a very well-known estimator so we will just describe the main stages of the algorithm step by step below:  
The state variable form of the states (in our case the coefficients) can be described as :
\begin{equation}
\texttt{x}_{k+1}=\texttt{F}_{k}\texttt{x}_{k}+\texttt{w}_{k}
\end{equation}
where, 
$\texttt{x}_{k}=\begin{bmatrix}
\texttt{x}_{1k}&\texttt{x}_{2k}&\texttt{x}_{3k}&\texttt{x}_{4k}&\texttt{x}_{5k}&\texttt{x}_{6k}&\texttt{x}_{7k}

\end{bmatrix}^{T}$ is the state vector  at $kth$ instant. The states are the coefficients of ${M}_{k}$ (i.e. $\texttt{x}_{1k}={e}_{1k},\texttt{x}_{2k}={e}_{2k},\texttt{x}_{3k}={e}_{3k},\texttt{x}_{4k}={m}_{1k},\texttt{x}_{5k}={m}_{2k},\texttt{x}_{6k}={s}_{1k},\texttt{x}_{7k}={s}_{2k}$).
 
$\texttt{F}_{k}={I}_{7}$ (identity matrix of size 7), is the state transition matrix

$\texttt{w}_{k}$ is process noise(zero mean white Gaussian type)

The observation signal can be represented using the following linear equation : 
\begin{equation}
\texttt{z}_{k}=\texttt{H}_{k}\texttt{x}_{k}+\texttt{v}_{k}
\label{eq_m}
\end{equation}

where, 
$\texttt{z}_{k}$ is the measurement vector at $kth$ instant and  $\texttt{v}_{k}$ is measurement noise(zero mean white Gaussian type).
 
$\texttt{H}_{k}$ is a vector which defines the relation between state and measurement vector. As our model treats original image as measurement signal and any expected block can be represented by only a single value, we will update the coefficients of $M_{k}$ using the mean value of all elements/pixels which belong to $\omega_{k}$ i.e.$\left \langle \texttt{w}_{k} \right \rangle$. So equation \ref{eq_m} can be rewritten as follows: 

\begin{equation}
\left \langle \texttt{z}_{k} \right \rangle=\texttt{H}_{k}\texttt{x}_{k}+\texttt{v}_{k}. 
\end{equation}
where, $ \texttt{z}_{k}= \omega_{k}$ and $\texttt{H}_{k}=\begin{bmatrix}
\texttt{h}_{1k}&\texttt{h}_{2k}&\texttt{h}_{3k}&\texttt{h}_{4k}&\texttt{h}_{5k}&\texttt{h}_{6k}&\texttt{h}_{7k}
\end{bmatrix}$ is a 1$\times$7 dimensional vector which contains the local statistics (for e.g  $\texttt{h}_{1k}=\text{Local entropy}_{scale_{1}}$).

Now the 7-state kalman filter can be expressed using the following equations: 
\begin{equation}
\hat{{\texttt{x}^{-}}_{k+1}}=\texttt{F}_{k}\hat{\texttt{x}_{k}}
\end{equation}
\begin{equation}
\texttt{P}^{-}_{k+1}=\texttt{F}_{k}\texttt{P}_{k}\texttt{F}^{T} _{k}+\texttt{Q}_{k}
\end{equation}
\begin{equation}
\texttt{K}_{k}=\texttt{P}^{-}_{k}\texttt{H}^{T}_{k}(\texttt{H}_{k}\texttt{P}^{-}_{k}\texttt{H}^{T}_{k}+\texttt{R}_{k} )^{-1}
\end{equation}
\begin{equation}
\hat{{\texttt{x}}_{k}}=\hat{{\texttt{x}^{-}}_{k}}+\texttt{K}_{k}(\texttt{z}_{k}-\texttt{H}_{k}\hat{{\texttt{x}^{-}}_{k}} )
\end{equation}
\begin{equation}
\texttt{P}_{k}=(\texttt{I}-\texttt{K}_{k}\texttt{H}_{k})\texttt{P}^{-}_{k}
\end{equation}
Where, $\texttt{Q}_{k}$ and $\texttt{R}_{k}$ are the process and measurement noise covariance corresponding to $\omega_{k}$. $\texttt{K}_{k}$ is the kalman gain and $\texttt{P}_{k}$ is the error covariance matrix. In our case the measure update equation of state of state vector has been slightly adjusted as shown below: 
\begin{equation}
\hat{{\texttt{x}}_{k}}=\hat{{\texttt{x}^{-}}_{k}}+\texttt{K}_{k}(\left \langle \texttt{z}_{k} \right \rangle-\texttt{H}_{k}\hat{{\texttt{x}^{-}}_{k}} )
\end{equation}
where $\texttt{z}_{k}={\omega}_{k}$

Now, as we have presented the key equations of the kalman filter it will be easy to see how we can simulate the method we have described in the previous section. The process noise covariance, $\texttt{Q}_{k}$ controls how much our estimated value will rely on the process (in our case prior belief) and measurement noise covariance, $\texttt{R}_{k}$ controls how much our prediction will be modulated by measurement. So if we choose high  $\texttt{Q}_{k}$ and low $\texttt{R}_{k}$, prediction will trust the measurements more and vice versa occurs when we choose low $\texttt{Q}_{k}$ and high $\texttt{R}_{k}$. So when error in prediction gets higher than a certain threshold value or we move between two blocks which are away from each other, we will increase  $\texttt{Q}_{k}$ and decrease $\texttt{R}_{k}$ and if both conditions are unsatisfied we will choose a low $\texttt{Q}_{k}$ and high $\texttt{R}_{k}$ . 

\begin{table}
\begin{center}
\begin{tabular}{|l|c|c|c|c|c|c}
\hline
$\texttt{P}_{0}$&$\texttt{x}_{0}$ & $\texttt{Q}_{1}$&$\texttt{R}_{1}$&$\texttt{Q}_{2}$&$\texttt{R}_{2}$\\
\hline\hline

${I}_{7}$& $\textbf{0}_{7\times1}$&$0.1\times$${I}_{7}$&${10}^{-10}$& ${10}^{ -10} \times{I}_{7}$&0.1\\

\hline
\end{tabular}
\end{center}
\caption{Kalman filter parameters used in our algorithm. $\texttt{P}_{0}$ and $\texttt{x}_{0}$ are respectivly intial error covariance matrix and intial state vector. $\texttt{Q}_{1}$ and $\texttt{R}_{1}$  belongs to the set-I of the noise covariance matrices and $\texttt{Q}_{2}$ and $\texttt{R}_{2}$ belongs to the second set. }
\label{table_p}
\end{table}
\subsection{Implementation details}
The first parameter we need to specify for our algorithm is the size of the blocks. To generate our results we have used blocks of size 25 x 25 (this has been selected empirically) and in all of our simulations, where we initially down-sampled the input image to 400$\times$300, this size provides satisfactory performance.

We already have shown that function ${M}_{k}$ incorporates three local statistics at different scales. To employ this, initially we calculated two local standard deviation maps (considering 3x3 and 5x5 neighborhood), two local mean maps (3x3 and 5x5 neighborhood) and three local entropy maps (5x5, 7x7 and 9x9 neighborhood) associated with the input image. Then when calculating ${M}_{k}$ for the kth block we simply taken mean of the values from this feature maps, over only the region which kth block specifies. Therefore the measurement vector $\texttt{H}_{k}$ contains the seven values corresponding to $\omega_{k}$ from these seven maps. 
\begin{figure}[t]
$
\begin{array}{ccccccccc}
\includegraphics[width=.7in]{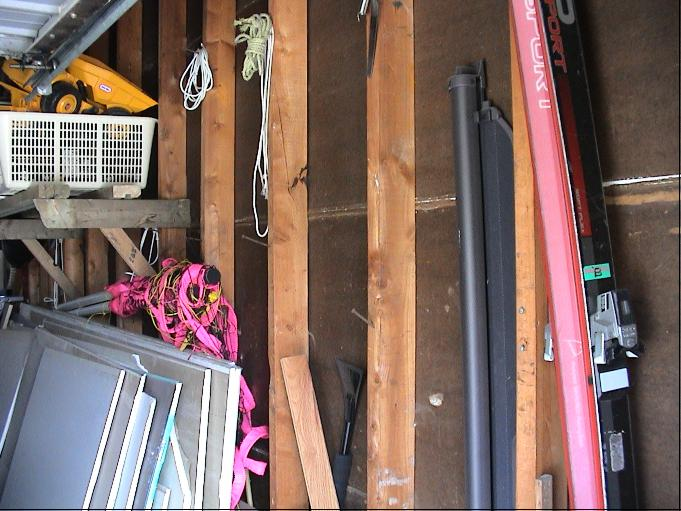}\hspace{-0.4em} &
\includegraphics[width=.7in]{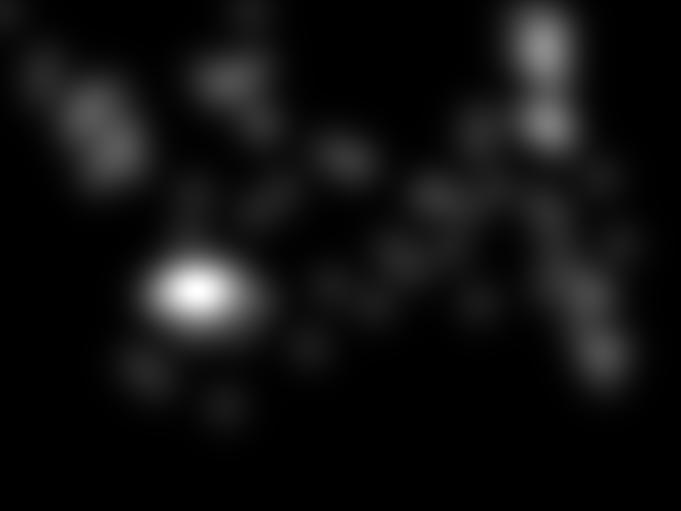}\hspace{-0.4em} &
\includegraphics[width=.7in]{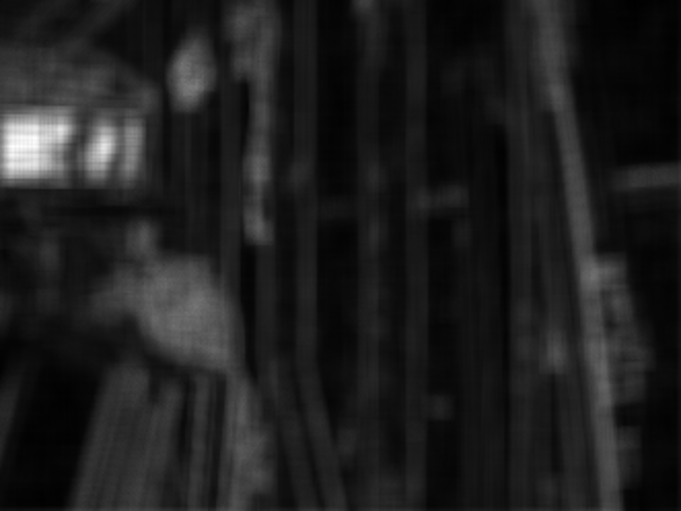}\hspace{-0.4em}&
\includegraphics[width=.7in]{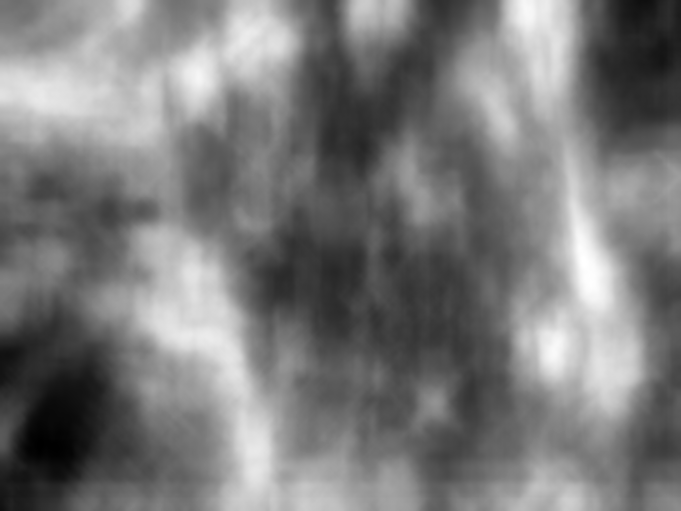}\hspace{-0.4em}&
\includegraphics[width=.7in]{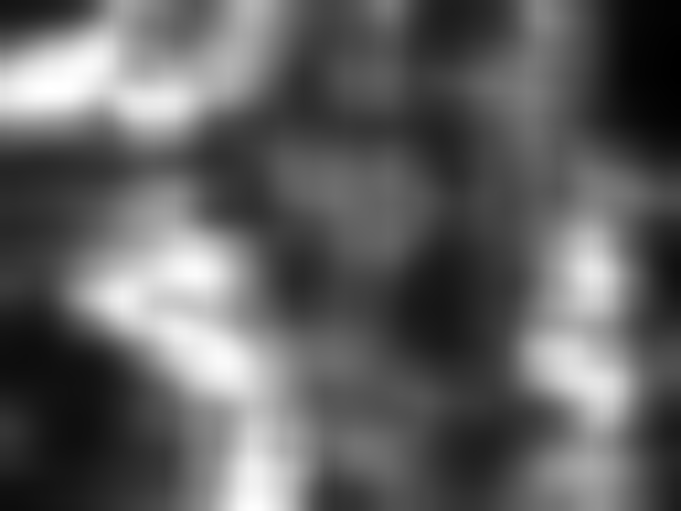}\hspace{-0.4em}&
\includegraphics[width=.7in]{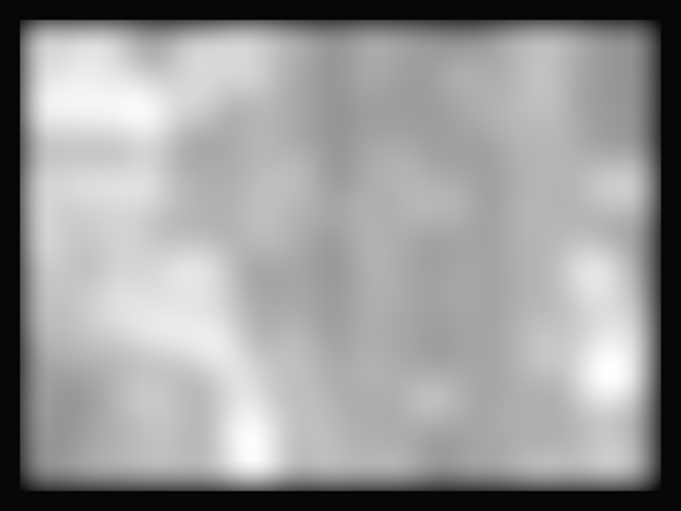}\hspace{-0.4em}&
\includegraphics[width=.7in]{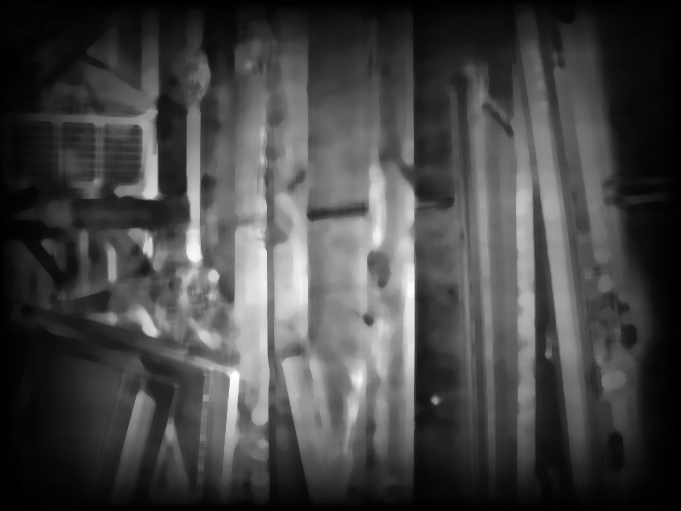}\hspace{-0.4em}&
\includegraphics[width=.7in]{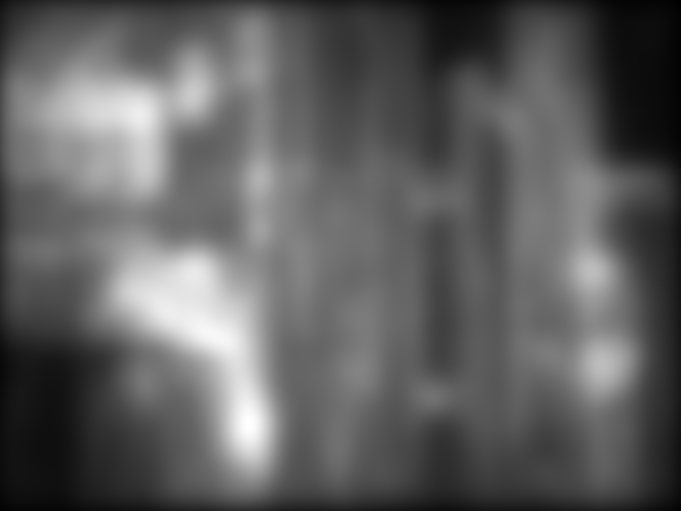}\\ 

\includegraphics[width=.7in]{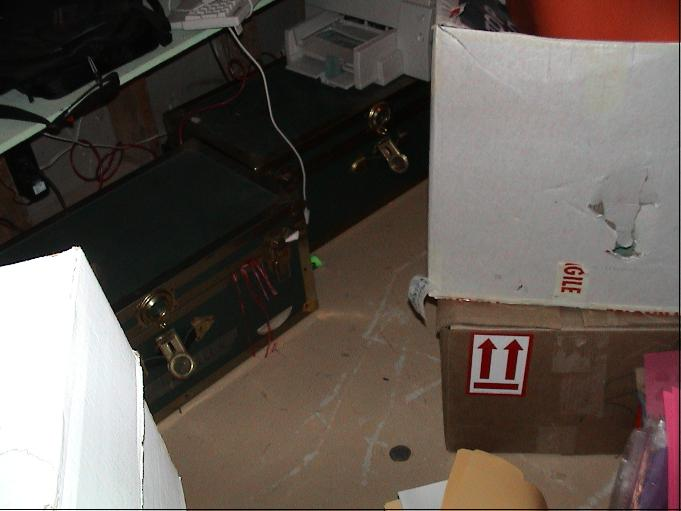}\hspace{-0.4em} &
\includegraphics[width=.7in]{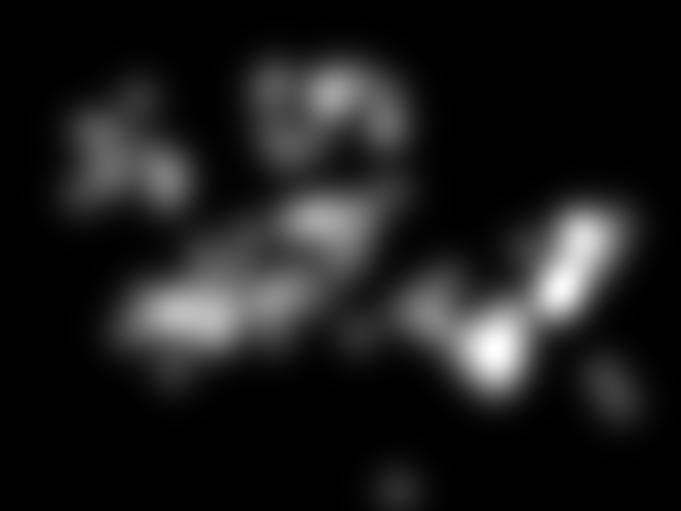}\hspace{-0.4em} &
\includegraphics[width=.7in]{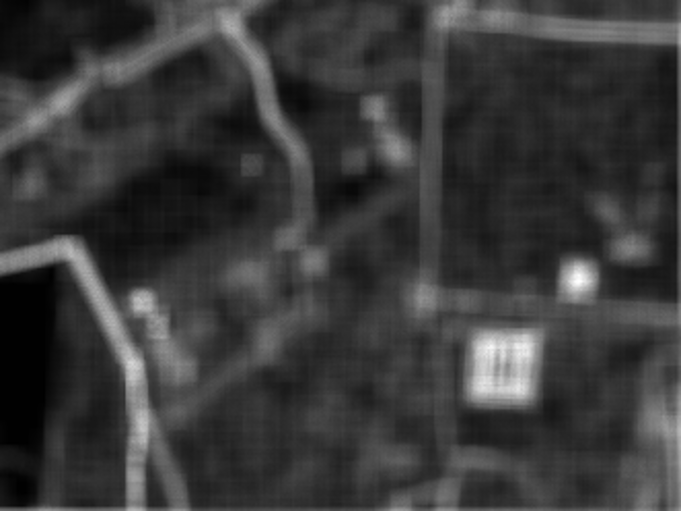}\hspace{-0.4em}&
\includegraphics[width=.7in]{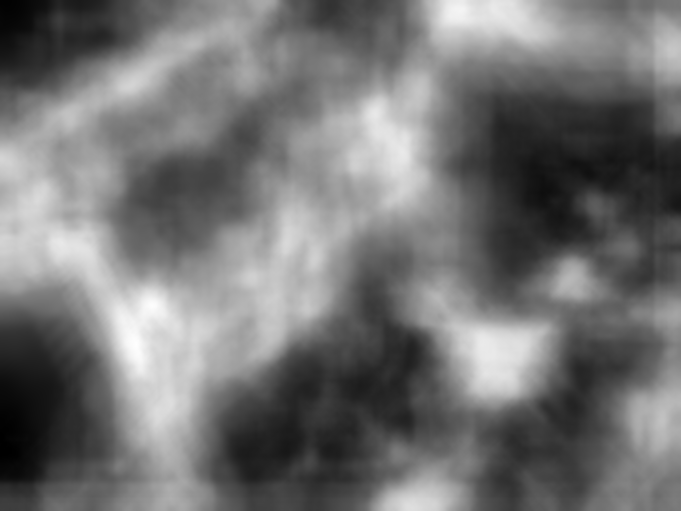}\hspace{-0.4em}&
\includegraphics[width=.7in]{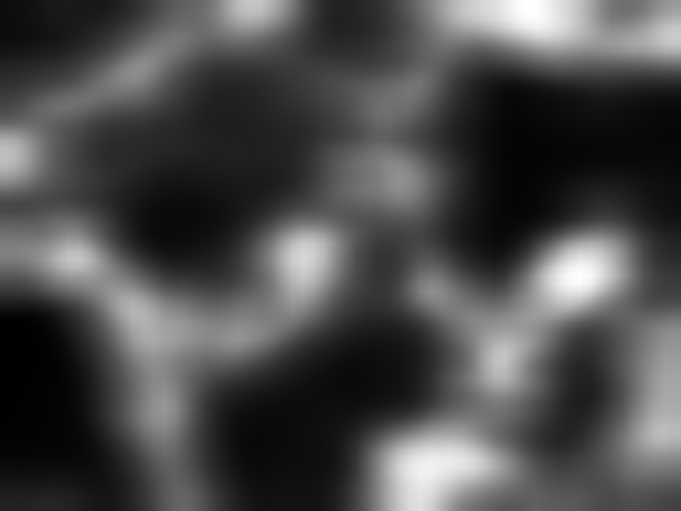}\hspace{-0.4em}&
\includegraphics[width=.7in]{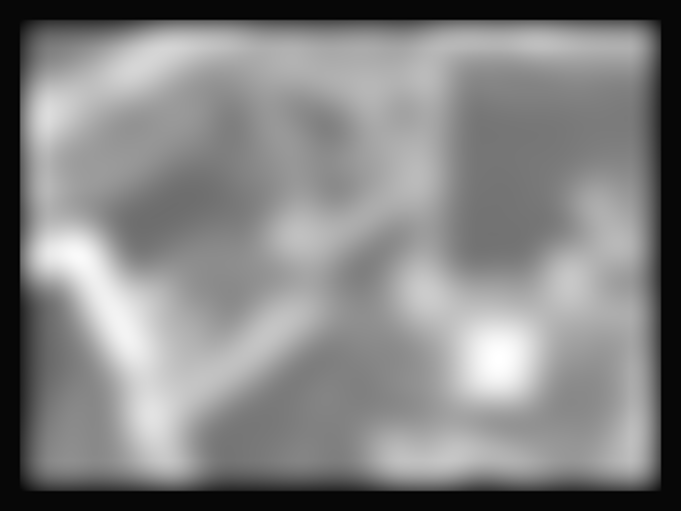}\hspace{-0.4em}&
\includegraphics[width=.7in]{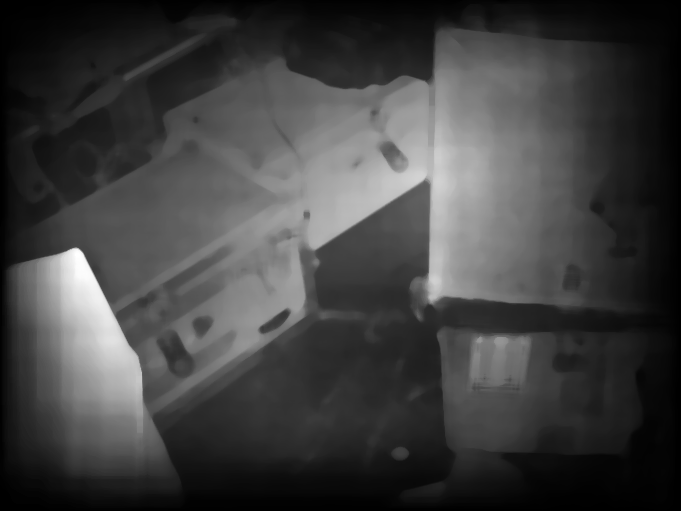}\hspace{-0.4em}&
\includegraphics[width=.7in]{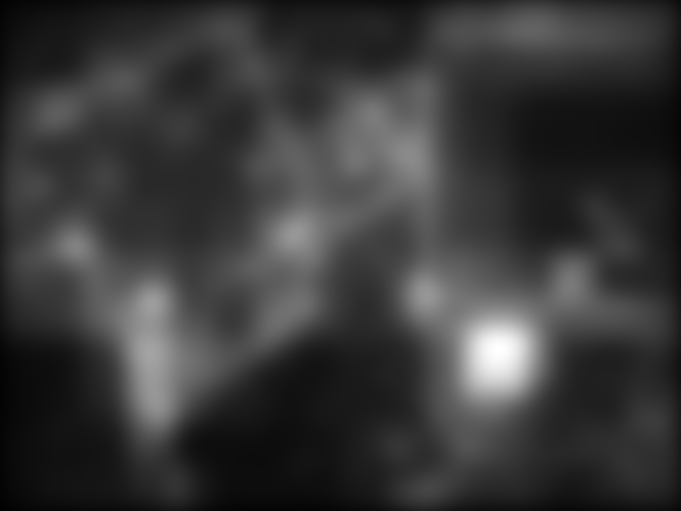}\\

\includegraphics[width=.7in]{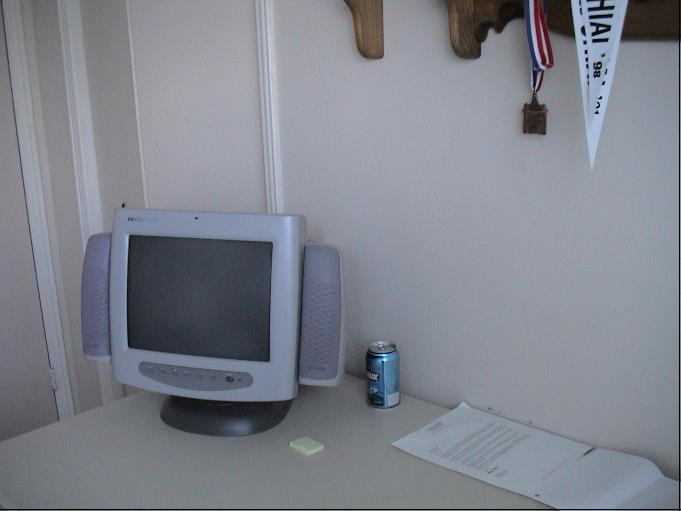}\hspace{-0.4em} &
\includegraphics[width=.7in]{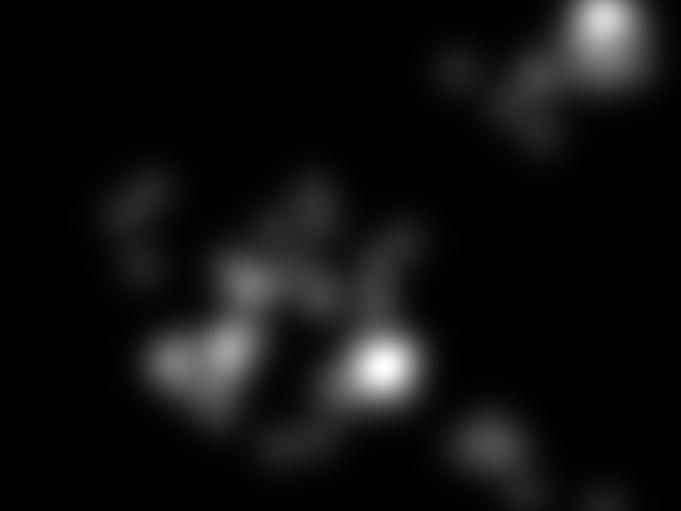}\hspace{-0.4em} &
\includegraphics[width=.7in]{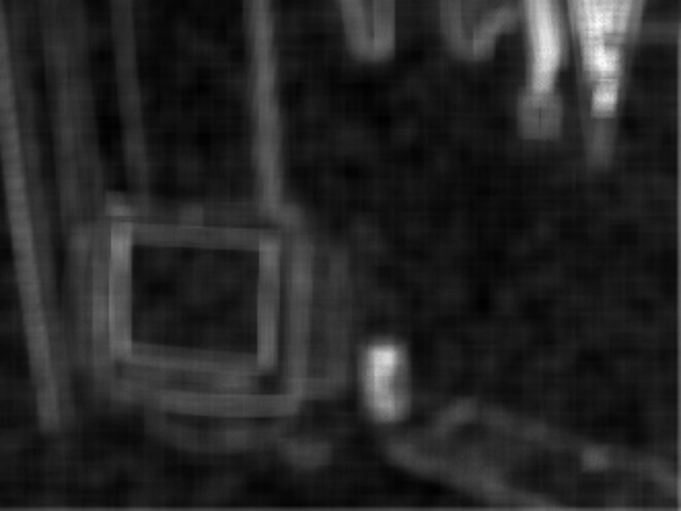}\hspace{-0.4em}&
\includegraphics[width=.7in]{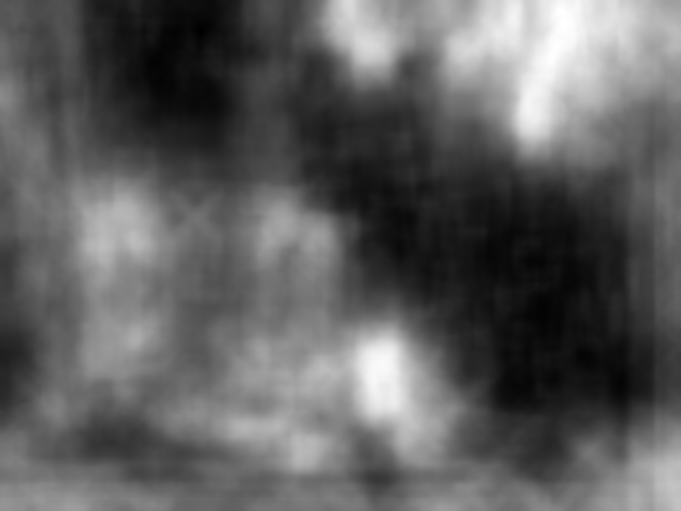}\hspace{-0.4em}&
\includegraphics[width=.7in]{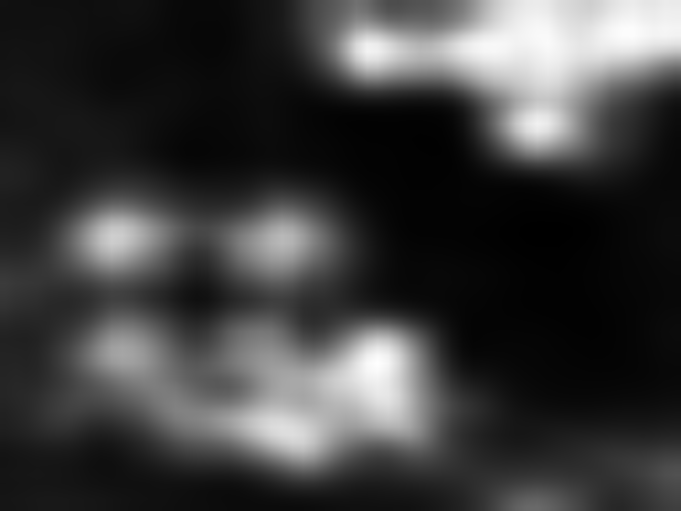}\hspace{-0.4em}&
\includegraphics[width=.7in]{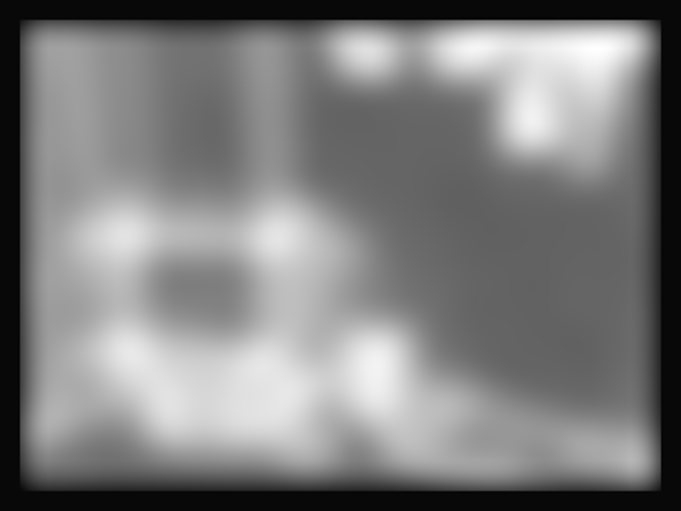}\hspace{-0.4em}&
\includegraphics[width=.7in]{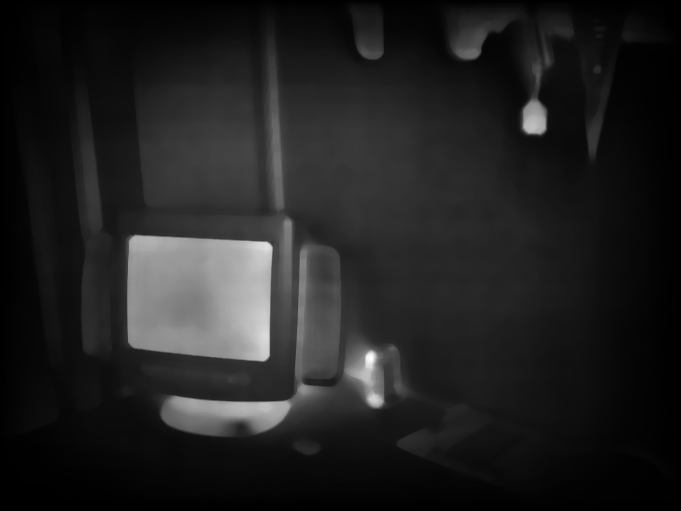}\hspace{-0.4em}&
\includegraphics[width=.7in]{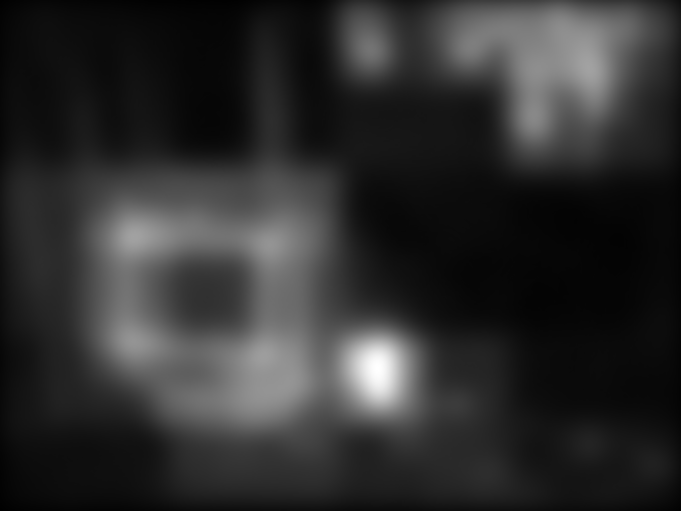}\\

\includegraphics[width=.7in]{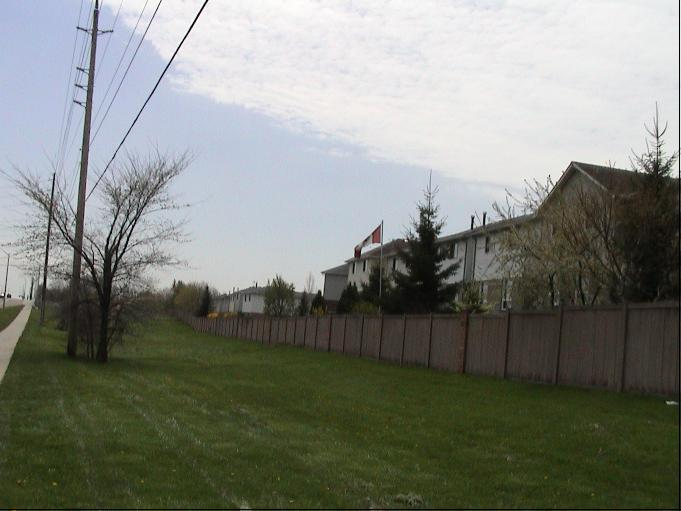}\hspace{-0.4em} &
\includegraphics[width=.7in]{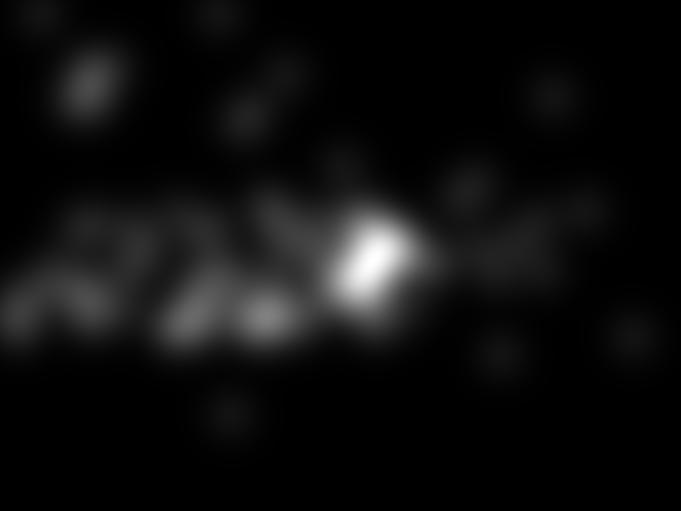}\hspace{-0.4em} &
\includegraphics[width=.7in]{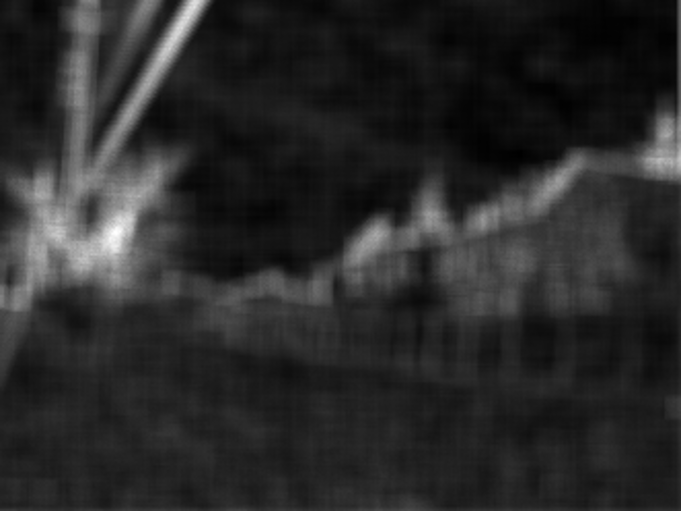}\hspace{-0.4em}&
\includegraphics[width=.7in]{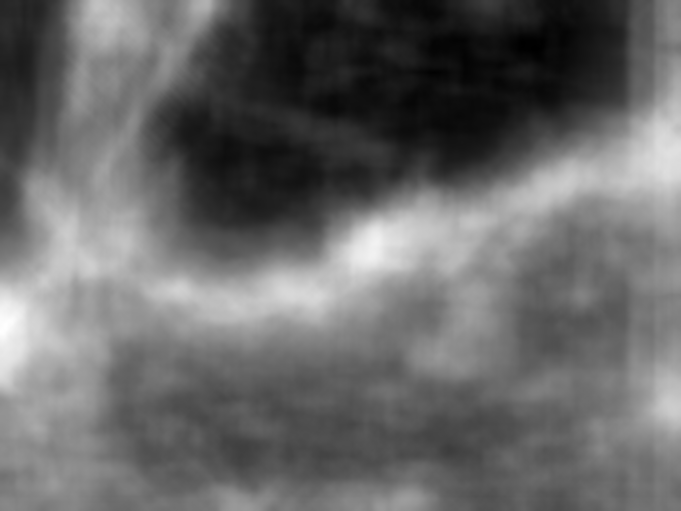}\hspace{-0.4em}&
\includegraphics[width=.7in]{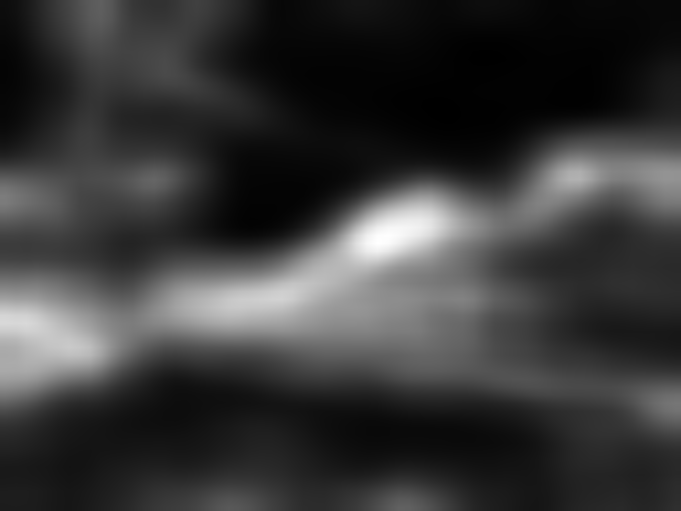}\hspace{-0.4em}&
\includegraphics[width=.7in]{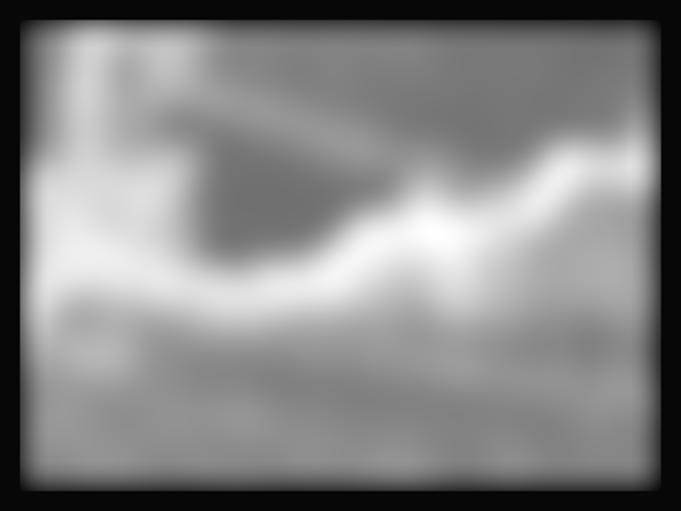}\hspace{-0.4em}&
\includegraphics[width=.7in]{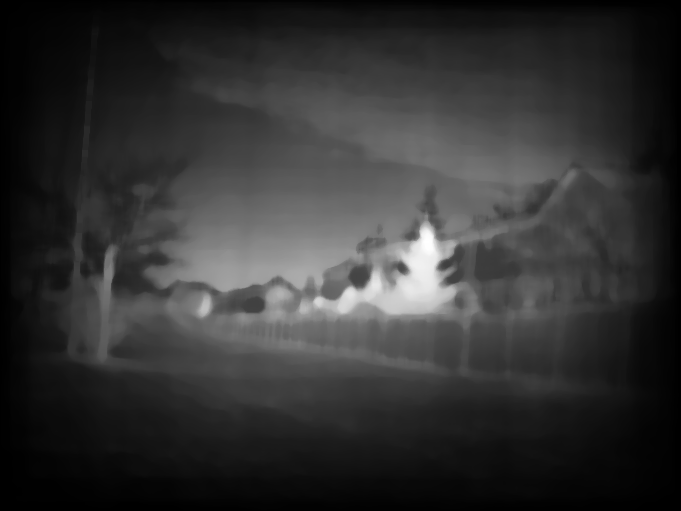}\hspace{-0.4em}&
\includegraphics[width=.7in]{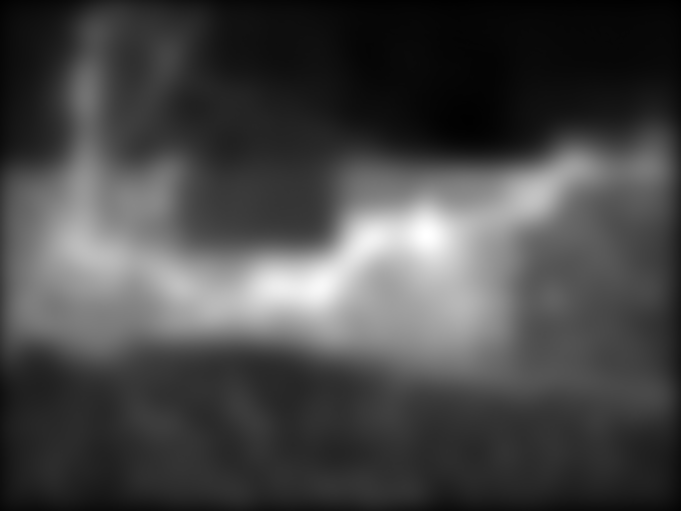}\\

\includegraphics[width=.7in]{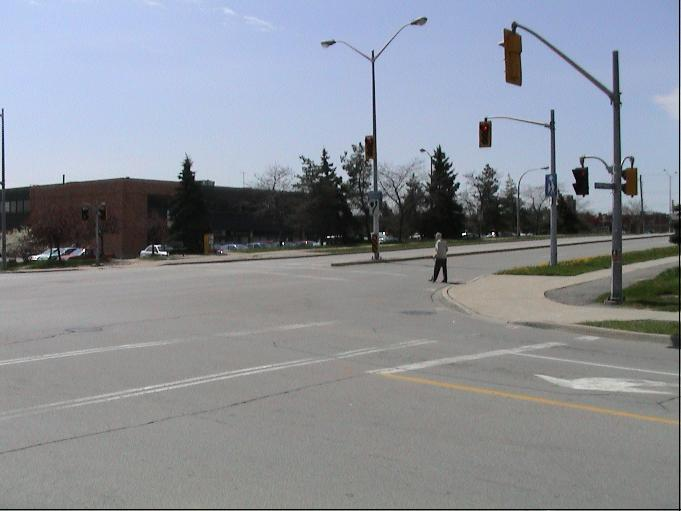}\hspace{-0.4em} &
\includegraphics[width=.7in]{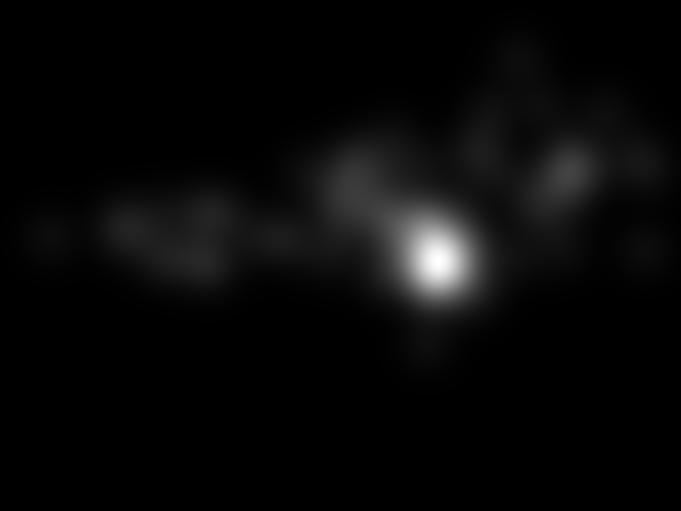}\hspace{-0.4em} &
\includegraphics[width=.7in]{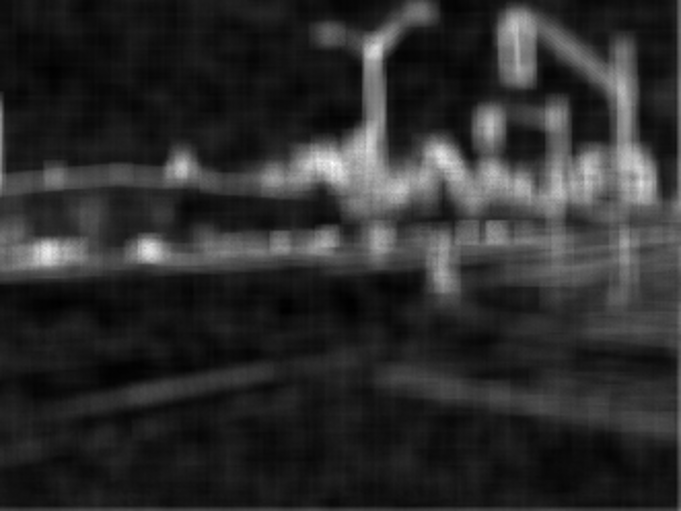}\hspace{-0.4em}&
\includegraphics[width=.7in]{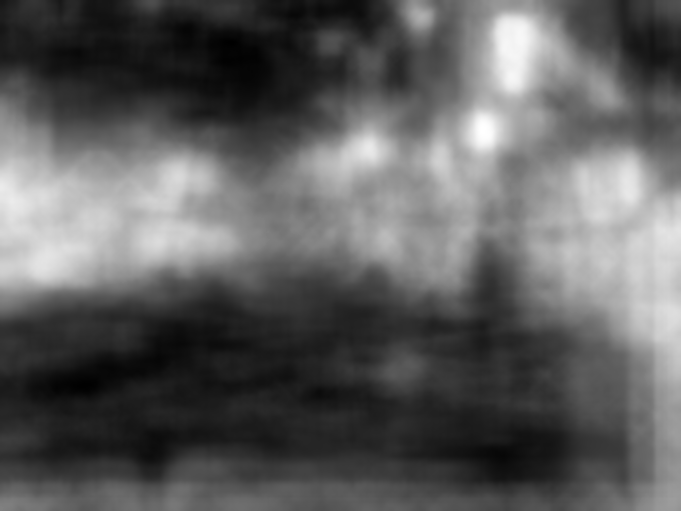}\hspace{-0.4em}&
\includegraphics[width=.7in]{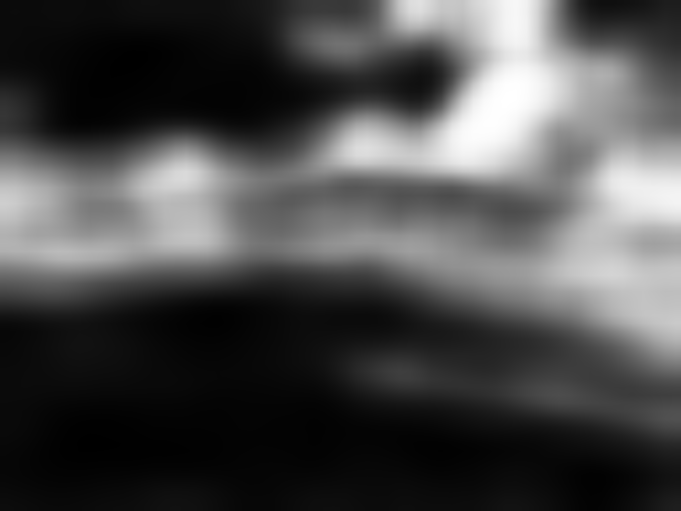}\hspace{-0.4em}&
\includegraphics[width=.7in]{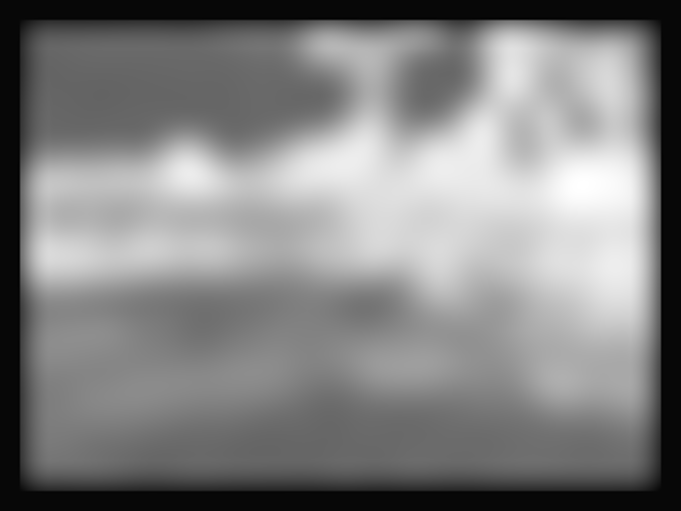}\hspace{-0.4em}&
\includegraphics[width=.7in]{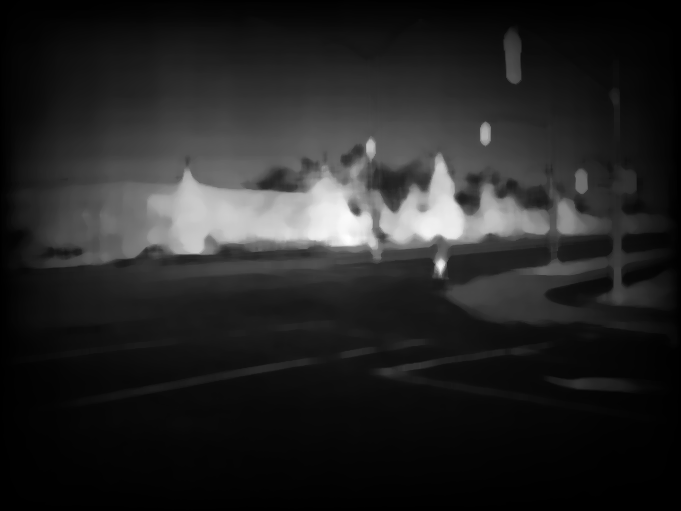}\hspace{-0.4em}&
\includegraphics[width=.7in]{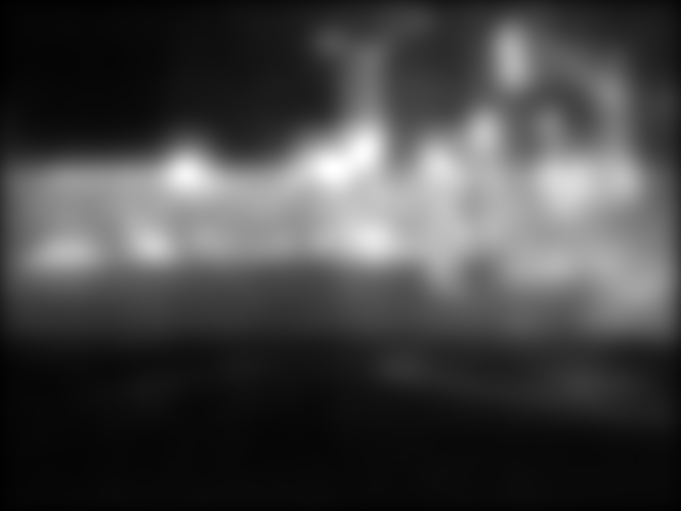}\\

\includegraphics[width=.7in]{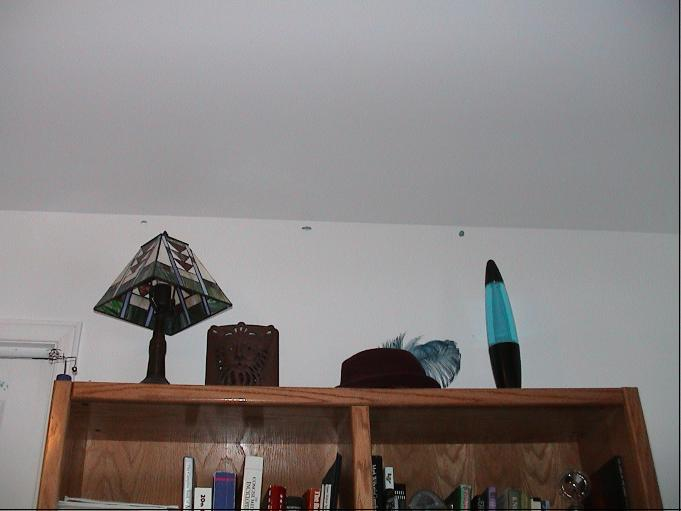}\hspace{-0.4em} &
\includegraphics[width=.7in]{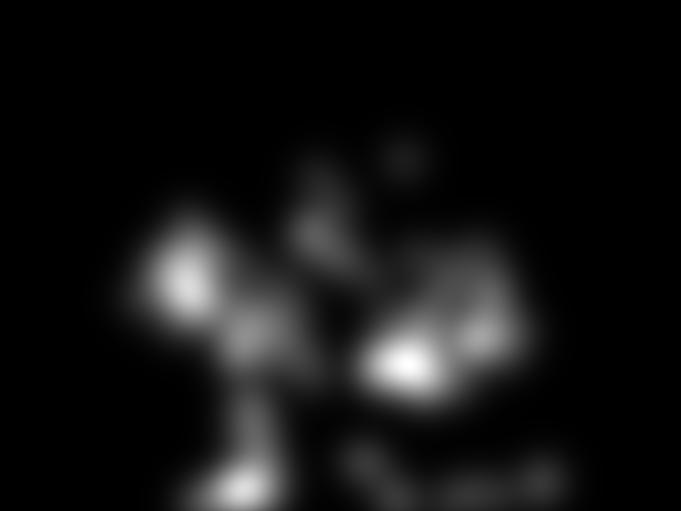}\hspace{-0.4em} &
\includegraphics[width=.7in]{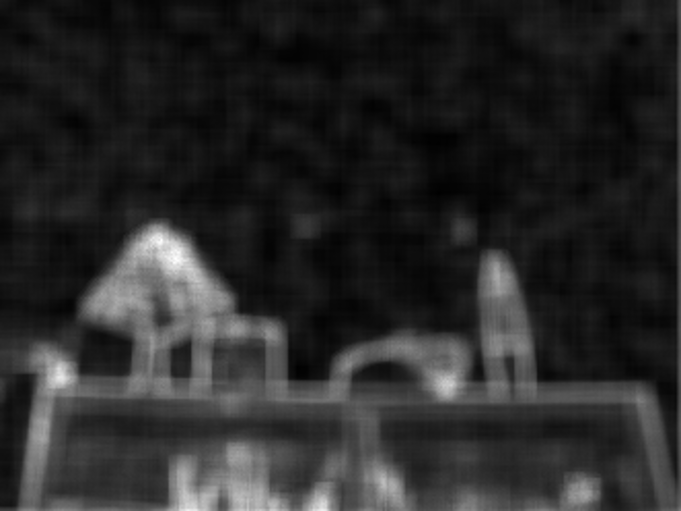}\hspace{-0.4em}&
\includegraphics[width=.7in]{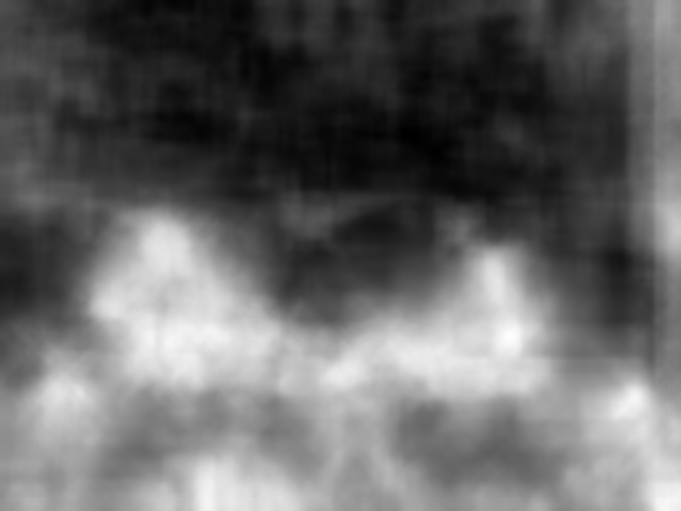}\hspace{-0.4em}&
\includegraphics[width=.7in]{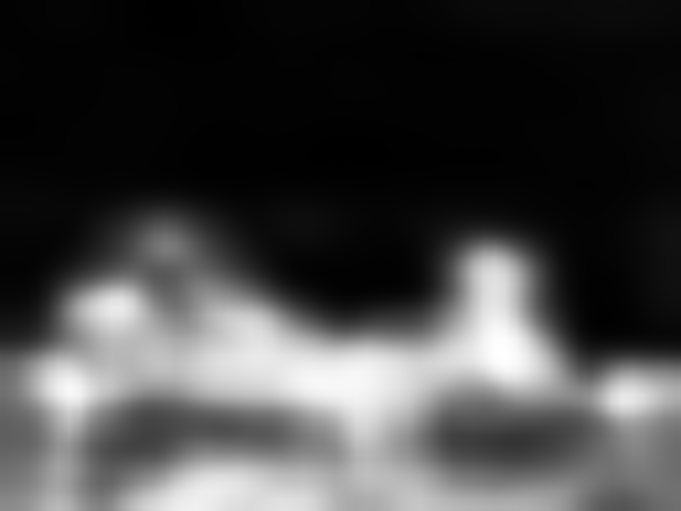}\hspace{-0.4em}&
\includegraphics[width=.7in]{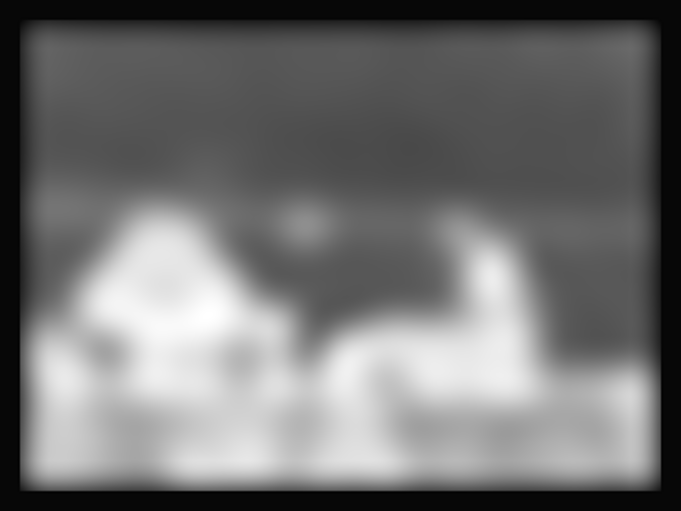}\hspace{-0.4em}&
\includegraphics[width=.7in]{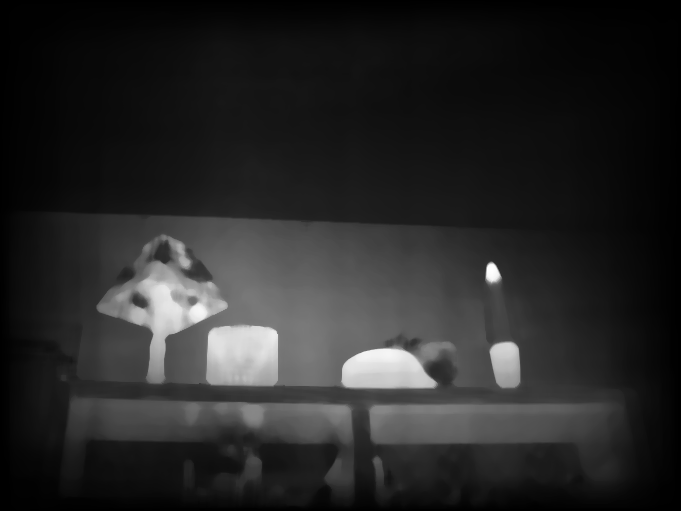}\hspace{-0.4em}&
\includegraphics[width=.7in]{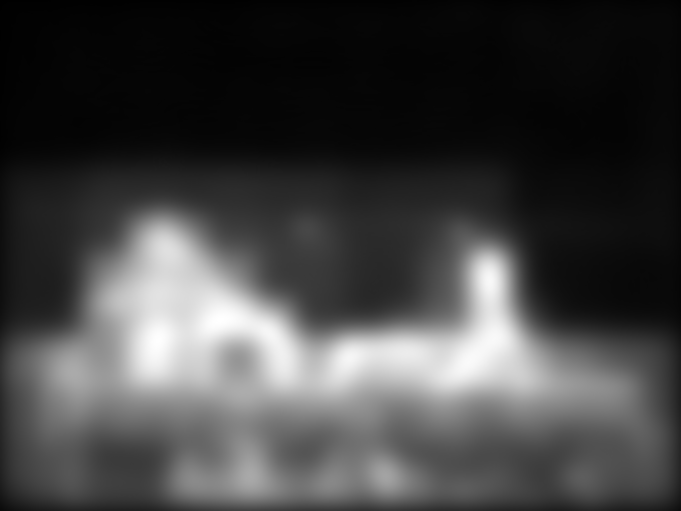}\\

\includegraphics[width=.7in]{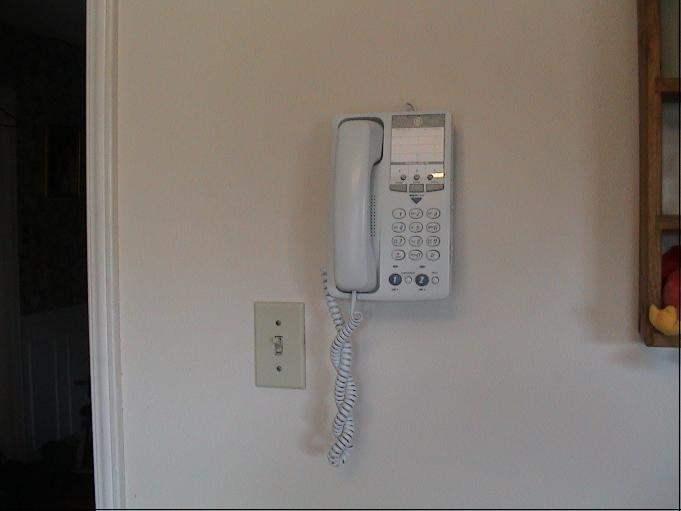}\hspace{-0.4em} &
\includegraphics[width=.7in]{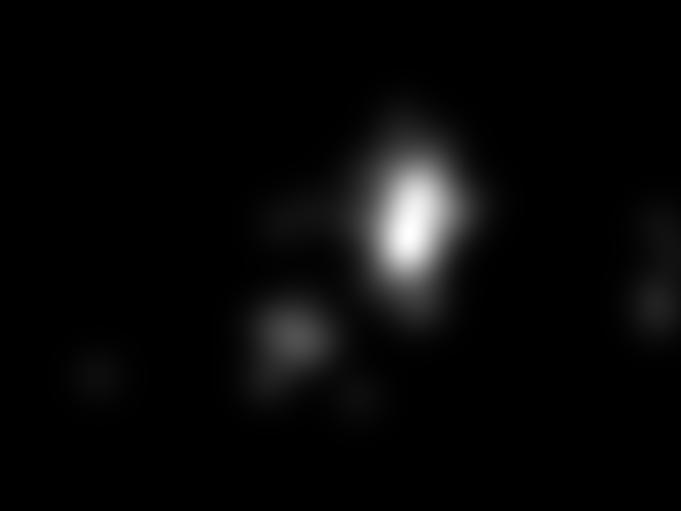}\hspace{-0.4em} &
\includegraphics[width=.7in]{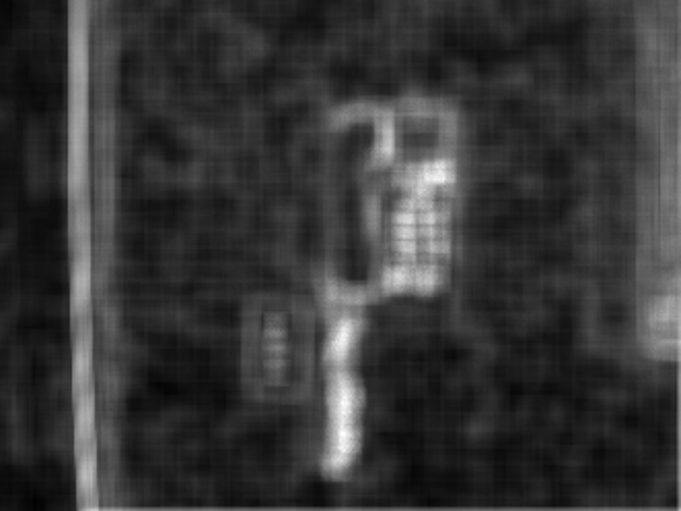}\hspace{-0.4em}&
\includegraphics[width=.7in]{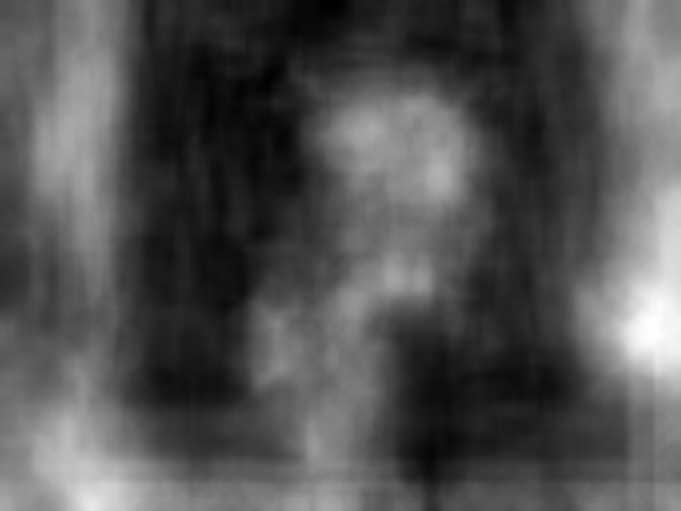}\hspace{-0.4em}&
\includegraphics[width=.7in]{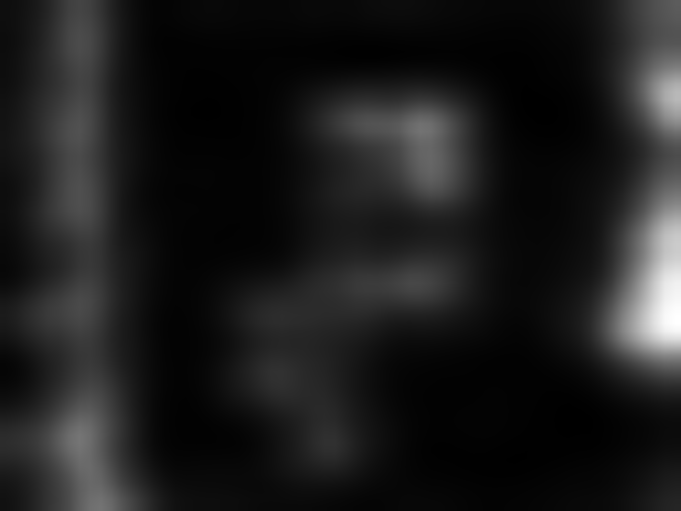}\hspace{-0.4em}&
\includegraphics[width=.7in]{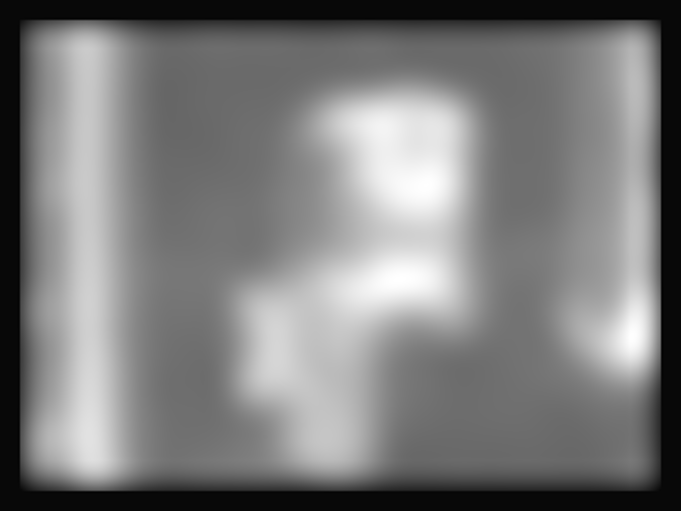}\hspace{-0.4em}&
\includegraphics[width=.7in]{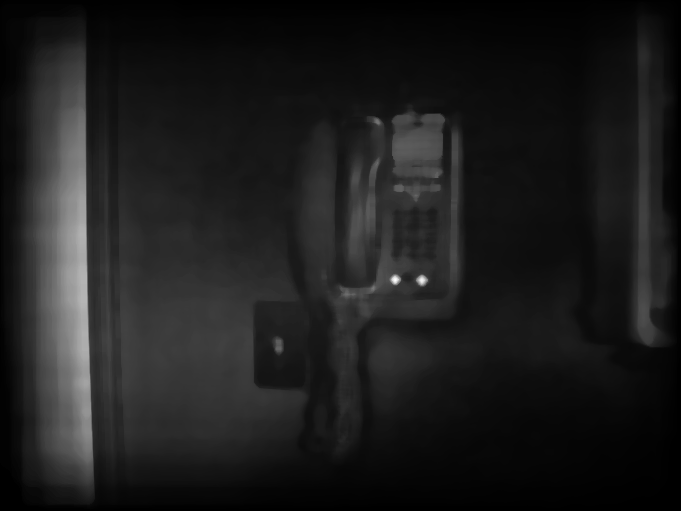}\hspace{-0.4em}&
\includegraphics[width=.7in]{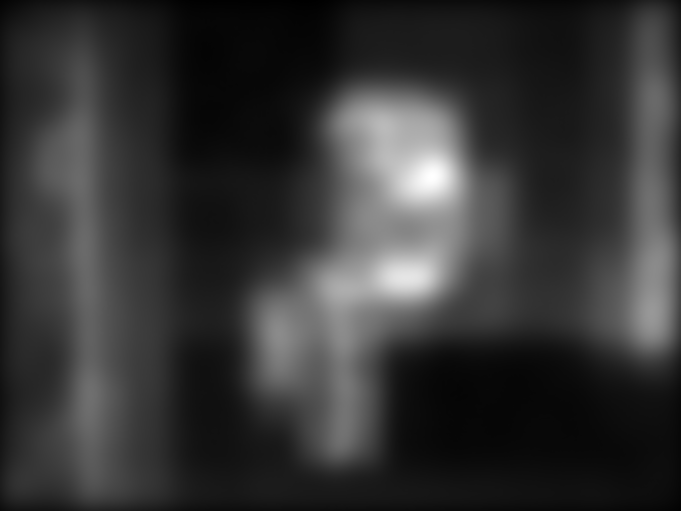}\\

\includegraphics[width=.7in]{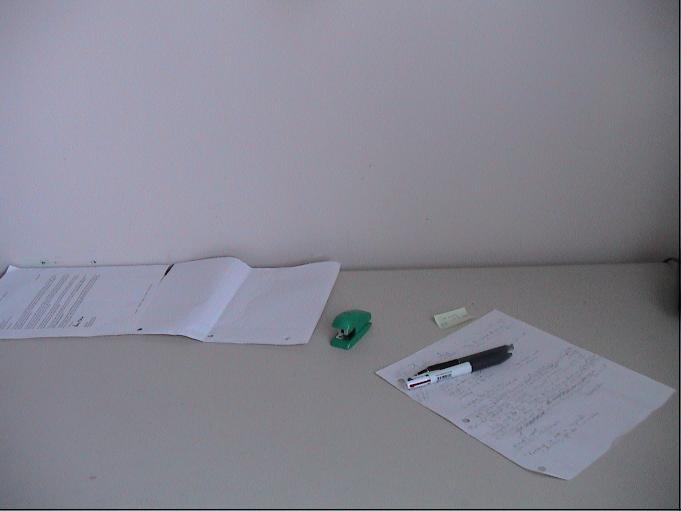}\hspace{-0.4em} &
\includegraphics[width=.7in]{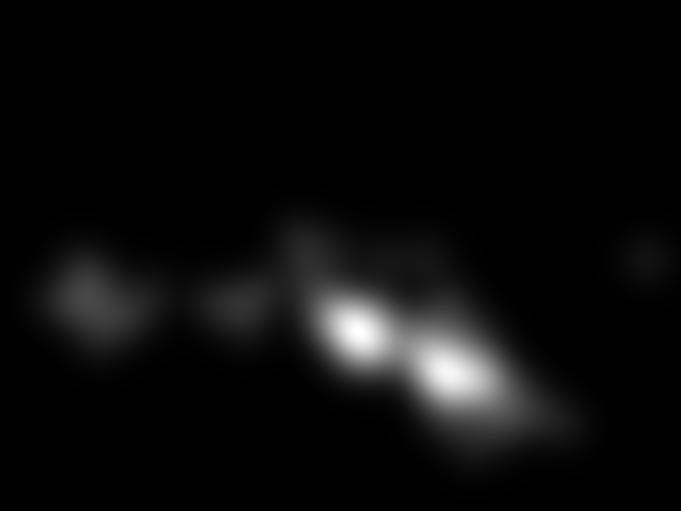}\hspace{-0.4em} &
\includegraphics[width=.7in]{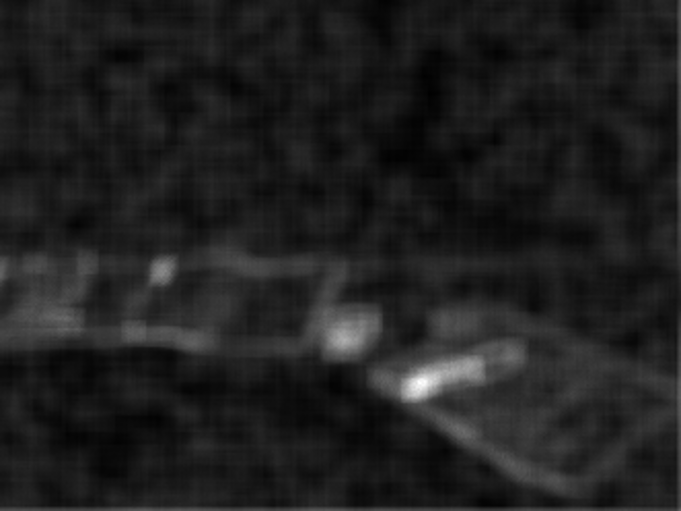}\hspace{-0.4em}&
\includegraphics[width=.7in]{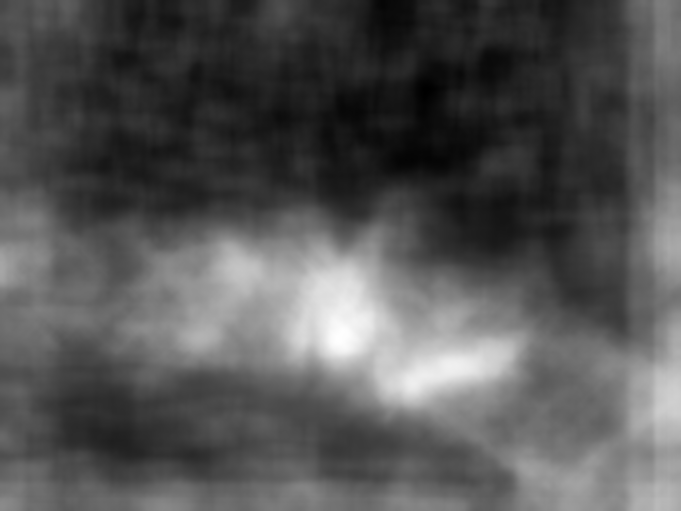}\hspace{-0.4em}&
\includegraphics[width=.7in]{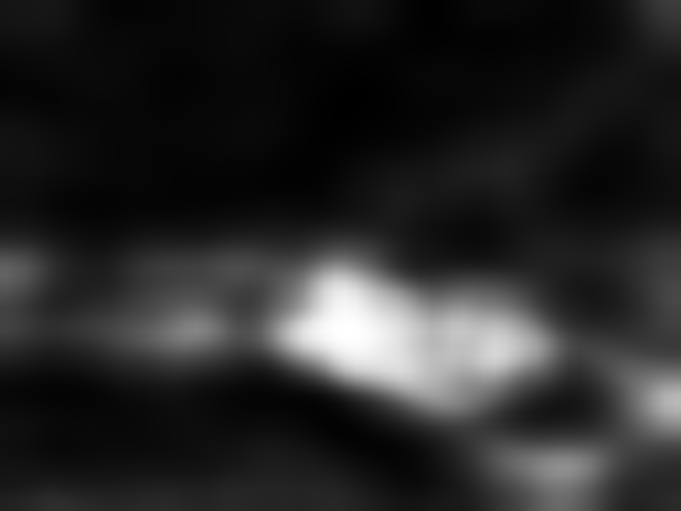}\hspace{-0.4em}&
\includegraphics[width=.7in]{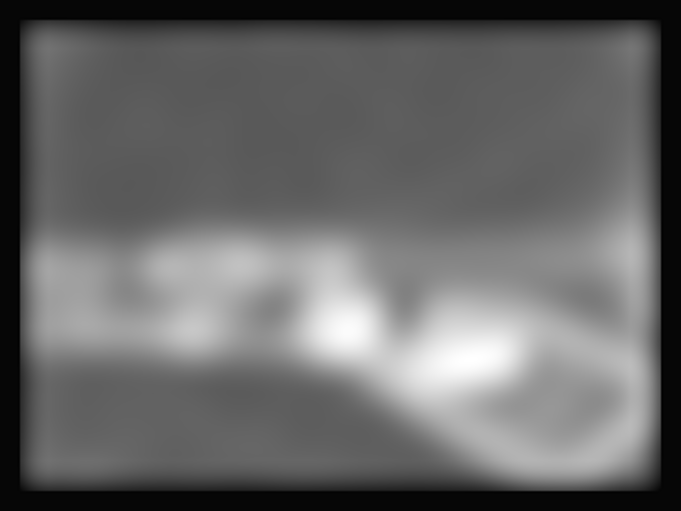}\hspace{-0.4em}&
\includegraphics[width=.7in]{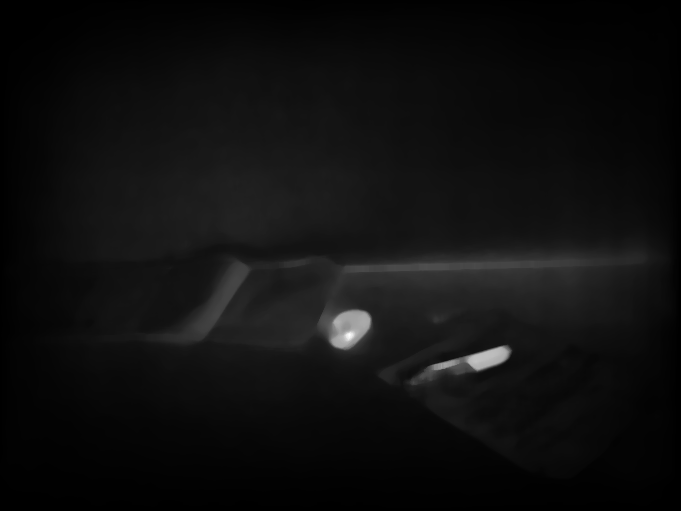}\hspace{-0.4em}&
\includegraphics[width=.7in]{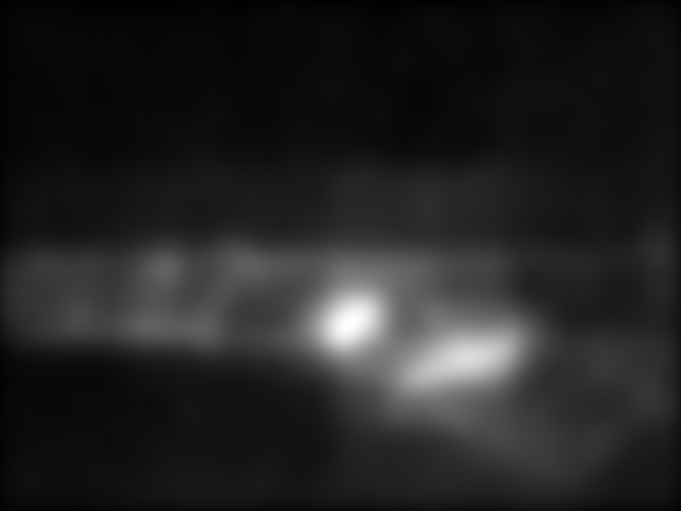}\\

\end{array}$

\vspace{-0.7em}
\caption{Sample results of our model along with human fixation density maps and results from other reference models. Column 1: Original input image, Column 2: Human fixation density map, Column 3: SUN~\cite{Zhang} , Column 4: CIWM~\cite{Murray}, Column 5: SR~\cite{Seo} ,Column 6: AIM~\cite{Bruce}, Column 7: RCSS~\cite{rcss}, Column 8: saliency maps from our proposed KS-7}
\label{visual_f}
\end{figure}
 Usually, tuning of kalman filter parameters is a challenging task but in our case tuning is not necessary as we are interested in coarse estimation of expected image. Furthermore, as expected image is a hypothetical signal and there is no precise definition of it, we cannot evaluate error in its estimation. Table \ref{table_p} contains the values of kalman filter parameters which we have used. As we have said earlier we will oscillate between two sets of values of $\texttt{R}_{k}$ and $\texttt{Q}_{k}$. When error between ${M}_{k}$ and $\left \langle \omega_{k} \right \rangle$ goes above a certain threshold error or we move from a block to another which does not belong to the neighborhood of the former (assuming 4-connectivity), we will use the values from set-I ($\texttt{Q}_{1}$ and $\texttt{R}_{1}$); otherwise we will use set-II ($\texttt{Q}_{2}$ and $\texttt{R}_{2}$). The error between  ${M}_{k}$ and $\omega_{k}$ can be defined as follows: 
\begin{equation}
Error_{k}=|M_{k}-\left \langle \omega_{k} \right \rangle|
\end{equation}

The random traversal strategy is not critical for our algorithm, we could traverse among the blocks in any manner. But if we move from one block to another only in a nearest neighbor(along any direction) sense, this can sometime slightly reduce the performance. Probabaly this is because continuously navigating along similar region can lead to almost no changes in coefficient values for a long time and then if a region even with little difference is introduced , it will produce more error. However in our algorithm we use large block size and avail a coarse construction strategy, so performance get negligibly affected by traversal strategy. 
Now if we don't move in nearest neighbor manner, distance between blocks will also decide how much our current expectation will be modulated by the prior belief. 

For multi-scale implementation, along with the initially scaled  input, we also produced saliency maps corresponding to the half and quarter resolution  images and then get the master saliency map combining these three saliency maps generated at three scales. 
\section{Experimental results}
We have evaluated our proposed algorithm against two benchmark datasets: 1) Toronto dataset~\cite{bruce2} and 2) MIT-300 dataset~\cite{juddreport}. The Toronto human fixation dataset, collected by Bruce and Tsotsos~\cite{bruce2}, is a well know benchmark dataset for visual saliency detection task  which contains 120 color images of equal dimensions (681 x 511) and eye fixation records from 20 human observers. MIT-300 is a relatively new dataset which contains 300 benchmark images of varying dimensions. It has been already stated that, in this work we have implented two different variants of our model. We will term the first implementation, which uses only three color channels, as "KS-3'' and the other one as "KS-7" which utilizes seven feature channels.
\begin{table}[h]
\begin{center}
\begin{tabular}{|l|c|c|c|c|c|c|c|}
\hline
 Model& AUC-Judd&AUC-Borji&CC&SIM&NSS\\
\hline\hline
AIM~\cite{Bruce} &0.79&0.77&0.33&0.38&0.83 \\
SUN~\cite{Zhang}&0.69&0.68&0.26&0.36&0.76 \\
SR~\cite{Seo}& 0.77&0.76&0.40&0.41&1.10\\
CIWM~\cite{Murray}&0 .75&0.74&0.35&0.38&0.96\\
RCSS~\cite{rcss}&0.78&0.76&0.43&\textbf{0.44}&1.16\\
KS-3 (proposed)& 0.79&0.78&0.44&0.42&1.20\\
KS-7 (proposed)& \textbf{0.83}&\textbf{0.82}&\textbf{0.53}&\textbf{0.44}&\textbf{1.42}\\
\hline
\end{tabular}
\label{table_r}
\end{center} 
\label{table_r}
\caption{Quantitative comparison between the proposed Kalman filter based method and other methods when predicting human eye fixations on Toronto data set~\cite{bruce2}. }
\end{table}

  \subsection{Evaluation metrics}
For quantitative evaluation we have used five standard metrics, namely AUC-Judd~\cite{juddreport}~\cite{riche}, AUC-Borji~\cite{riche}, Correlation Coefficient (CC)~\cite{riche}, Similarity measure (SIM)~\cite{juddreport}~\cite{riche} and Normalized Scanpath Saliency (NSS)~\cite{nss}~\cite{riche}. AUC-Judd and AUC-Borji are both area under the ROC(Receiver operating characteristic) curve based metrics which convert the saliency maps into binary maps and treat them as classifiers. 
The third metric CC or correlation coefficient is a linear measure which can be calculated as below: 

\begin{equation}
CC=\frac{cov(S,F)}{\sigma _{S}\ast \sigma _{F}}
\end{equation}
where $S$ and $F$ are saliency map and human fixation map respectively.

The output range of CC metric is between -1 to +1. $|\text{CC output}| = 1$ denotes there is perfect linear relationship exists between the ground fixation density map and saliency map.  The similarity metric (SIM) compares fixation map and saliency map when they are viewed as normalized distributions. The similarity measure between normalized distributions,$S_{n}$  and $F_{n}$ can be given by: 

\begin{equation}
SIM= \sum_{x=1}^{N}min(S_{n}(x),F_{n}(x))
\end{equation}
where, $\sum_{x=1}^{N}S_{n}(x)=1$ and  $\sum_{x=1}^{N}F_{n}(x)=1$

A similarity score of 1 denotes that the two distributions are identical. The last metric we used for evaluation is Normalized Scanpath Saliency or NSS. This quantitative measure was proposed by Peteers and Itti ~\cite{nss} in 2005. The overall NSS score of a saliency map  can be given by: 
\begin{equation}
NSS=\frac{1}{N}\sum_{x=1}^{N}S_{n}(x)
\end{equation}
where $S_{n}$ and $N$ denote normalized saliency map and the total number of human eye fixations respectively. 

  \subsection{Performance on Toronto data set}
On Toronto dataset,  We have compared our results with five other saliency models which are:  1) SUN~\cite{Zhang} 2) Information maximization model(AIM)~\cite{Bruce} 3) Self resemblance model(SR)~\cite{Seo} and 4) Chromatic Induction wavelet model(CIWM)~\cite{Murray} and 5) Random Center Surround Model(RCSS)~\cite{rcss}. Performances have been compared with the other methods both quantitatively (Table 2) and visually (Fig.\ref{visual_f}). 
From Table 2, we can easily see that the 7 channel variant(KS-7) of the proposed approach outperformed the other models against all metrics by a wide margin. Only Random Center Surround Saliency model gave similar performance in terms of similarity score. The 3 channel variant of our algorithm, KS-3 also achieved state of the art performance on all metrics. As we can see from the example images, our method is less susceptible to the edges than Zhang et al(SUN) and Bruce\& Tsotsos(AIM). Though, CIWM and  Self resemblance model(SR) sometimes demonstrated better edge suppression, these models tended to include large non-salient regions. When contrast between salient region and background is relatively low (e.g. sample image 7 in Figure 3), only Bruce-Tsotos and Our model performed well. Zhang's method(SUN) mainly highlighted edges for most of the images.From visual inspection, it seems that RCSS model is more appropriate for salient object detection rather than eye fixation prediction. RCSS also gave very poor performance for both high entropy images and low contrast images. Qualitative comparison among results from different models also suggests that a large part of our success can be attributed to the significant reduction in false detection. 

In Figure 4, we have demonstrated the ROC (Receiver operating characteristic) curves for CIWM~\cite{Murray}, RCSS~\cite{rcss} and the proposed method. As we can see from the plots, KS-7 demonstrates greater efficacy than the other models.  ROC  curves of RCSS and KS-3 are close to each other while performance of CIWM is inferior to the other 3 models. 
\begin{figure}[t]
\begin{center}
\includegraphics[width=3 in,height=2.8in]
                   {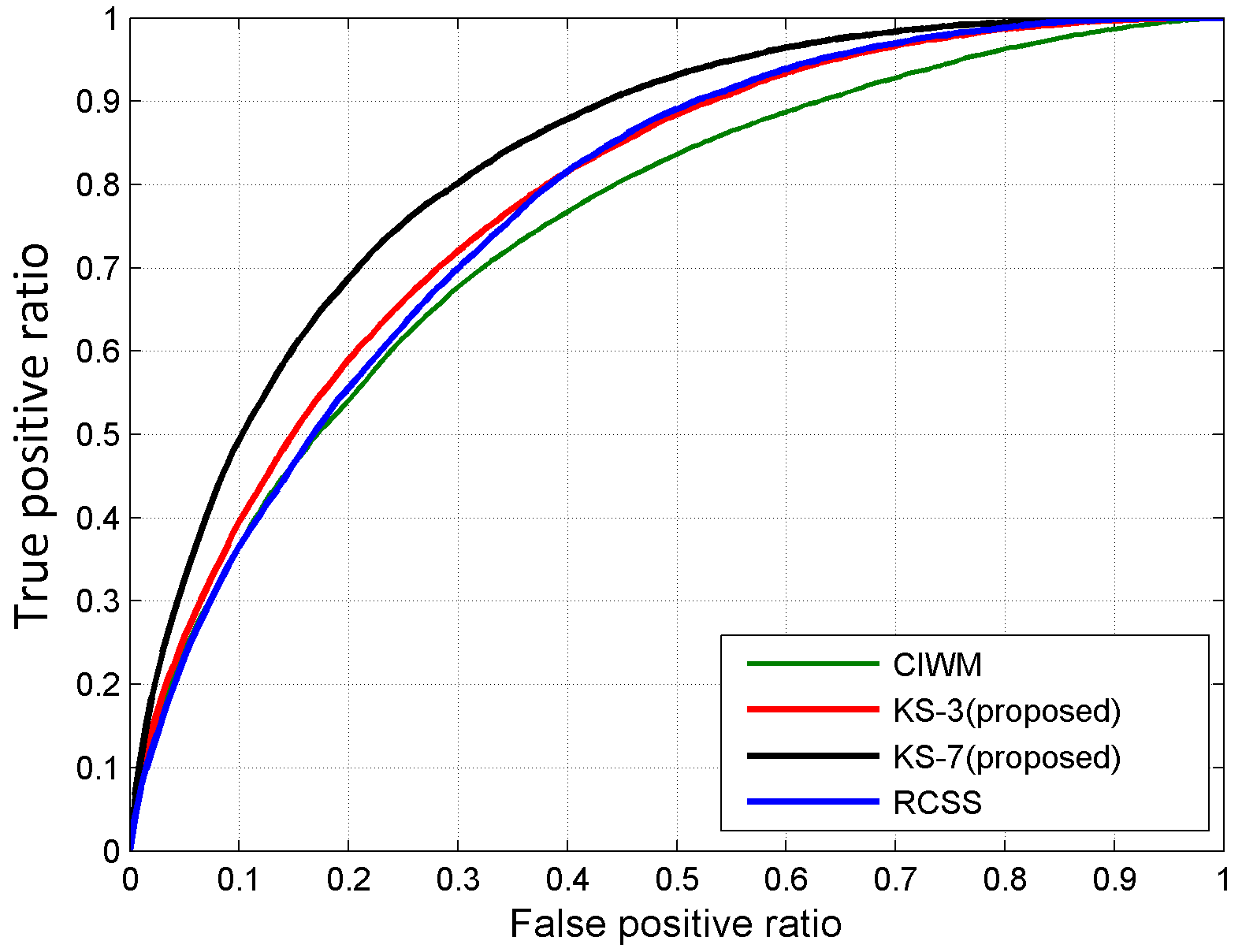}
\end{center}
   \caption{ROC curves from both of the versions of our model, CIWM~\cite{Murray} and RCSS model~\cite{rcss}}
\label{fig:short}
\end{figure}
  \subsection{Performance on MIT-300 data set}
As the ground fixation maps for MIT-300 images are not publicly available, we compared our model only quantitatively with the other approaches on this dataset. In addition to the 5 models used for comparison on Toronto dataset, we have assessed our model’s performance on MIT-300 against   7 other state of the art methods which are: CNN-VLM~\cite{cnn}, Multiple Kernel Learning model (MKL)~\cite{mkl}, Context Aware Saliency (CAS)~\cite{context}, Generalized Nonlocal Mean Saliency (GNMS)~\cite{gnms}, NARFI saliency (NARFI)~\cite{narfi}, Sampled Template Collation (STC)~\cite{stc} and LGS model~\cite{lgs}. In table 3, we have presented quantitative performance of various models on MIT-300 data set and these results from MIT-300 clearly demonstrates the superiority of our kalman based method which  outperformed all other approaches against AUC-Judd, AUC-Borji and CC metric. On SIM and NSS metric also the proposed approach(KS-7) achieved top scores along with RCSS and CNN-VLM model. Despite being a completely low level model our method performed better in an overall manner than the two learning based approaches: CNN-VLM and MKL. The proposed approach also gave significantly better saliency predictions than Context Aware Saliency model(CAS) which uses higher level feature detectors (such as face detector) as well as low level detectors.
\begin{table}[h]
\begin{center}
\begin{tabular}{|l|c|c|c|c|c|c|c|}
\hline
 Model& AUC-Judd&AUC-Borji&CC&SIM&NSS\\
\hline\hline
CNN-VLM~\cite{cnn}& 0.79&\textbf{0.79}&0.44&0.43&\textbf{1.18} \\
MKL~\cite{mkl}& 0.78&0.78&0.42&0.42&1.08 \\
CAS~\cite{context}& 0.74&0.73&0.36&0.43&0.95\\
LGS~\cite{lgs}& 0.76&0.76&0.39&0.42&1.02\\
GNM~\cite{gnms}& 0.74&0.67&0.34&0.42&0.97\\
NARFI~\cite{narfi} & 0.73&0.61&0.31&0.38&0.83\\
STC~\cite{stc}& 0.79&0.78&0.40&0.39&0.97 \\
RCSS~\cite{rcss}& 0.75&0.74&0.38&\textbf{0.44}&0.95\\
CIWM~\cite{Murray} & 0.70&0.69&0.27&0.38&0.73 \\
SUN~\cite{Zhang}&0.67&0.66&0.25&0.38&0.68\\
AIM~\cite{Bruce}&0.77&0.75&0.31&0.40&0.79\\
SR~\cite{Seo}&0.71&0.69&0.31&0.41&0.83\\
KS-7 (proposed)&\textbf{ 0.80}&\textbf{0.79}&\textbf{0.46}&\textbf{0.44}&\textbf{1.18} \\

\hline
\end{tabular}
\label{table_r}
\end{center} 
\caption{Quantitative comparison between the proposed Kalman filter based method and other methods when predicting human eye fixations on MIT-300 data set~\cite{juddreport}. }
\end{table}

\section{Conclusion}
In this paper we have presented a Kalman filter based  saliency detection method which generates a “visually expected scene” and based on that builds a saliency map. We have developed our model around the notion of “visual surprise” and it can be extended easily for video data, where instead of traversing the spatial domain, we will progress through the time domain. Our proposed model also provides a great deal of flexibility as anybody can use their own definition of the function, ${M}_{k}$, combining multiple features. We have evaluated two different implementations of our model using the two popular benchmark data sets and compared our results with various other established algorithms. Experiments have showed that the proposed model performs considerably better than several other existing methods. 

In future we would like explore pre-attentive segmentation models for initial segmentation instead of just dividing into blocks of uniform size in an ad hoc manner. Optimization of model implementation also demands detailed investigation as we did not attempt it yet. Finally, we would like to examine if performance can be improved by means of introducing non-linear combination of local statistics instead of linear one.

{

\end{document}